# Block Expanded DINORET: Adapting Natural Domain Foundation Models for Retinal Imaging Without Catastrophic Forgetting

J. Zoellin[1,2#], C. Merk[1,2#], M. Buob[1,2#], A. Saad[1,2], S. Giesser[1,2], T. Spitznagel[1,2], F. Turgut[1,2,3], , R. Santos[4], Y. Zhou[5,6,7], S. Wagner[5,6], Y.C. Tahm[8,9,10], P. A. Keane[5,6], D. Cabrera DeBuc[11,12, *], M. D. Becker[1,2,13,*], G. M. Somfai[1,2,14, *]

[#]These authors contributed equally as shared-first authors

[*] These authors contributed equally as senior and corresponding authors for this work

[1]Department of Ophthalmology, Stadtspital Zürich, Zurich, Switzerland,

[2]Spross Research Institute, Zurich, Switzerland,

[3] Gutblick Research, Switzerland,

[4]Digital Medicine Unit, Balgrist University Hospital, Zurich, Switzerland,

[5]NIHR Biomedical Research Centre at Moorfields Eye Hospital NHS Foundation Trust, London, UK,

[6]Institute of Ophthalmology, University College London, London, UK,

[7]Department of Medical Physics and Biomedical Engineering, University College London, London,

[8]Centre for Innovation and Precision Eye Health, National University of Singapore, Singapore,

[9]Department of Ophthalmology, National University of Singapore, Singapore,

[10]Singapore Eye Research Institute, Singapore National Eye Centre, Singapore,

[11]Ophthalmology & Visual Sciences Academic Clinical Program, Duke-NUS Medical School, Singapore,

[11]iScreen 2 Prevent LLC, Miami, FL, USA,

[12]Bascom Palmer Eye Institute, University of Miami, FL, USA ,

[13]Department of Ophthalmology, University of Heidelberg, Heidelberg, Germany,

[14]Department of Ophthalmology, Semmelweis University, Budapest, Hungary

**Disclosures:** The Authors declare no relevant financial interests in the conducted work.

**Corresponding Authors:**

**Dr. Gábor Márk Somfai, PhD**

Department of Ophthalmology,   Stadtspital Zurich,   Birmensdorferstrasse 497,   CH-8063 Zürich

Phone: +41 44 416 11 11

Fax +41 44 416 26 00

e-mail: somfaigm@yahoo.com

**Delia Cabrera DeBuc**

iScreen 2 Prevent LLC (DBA Multinostics)

9511 SW 6th CT, Pembroke Pines, FL, USA 33025

Phone: 954-391-9716

Email: delia_debuc@iscreen2prevent.com

Keywords: Foundation Model, Computer Vision, Domain Adaptation, DINOv2, Ophthalmology, Catastrophic Forgetting, Block Expansion, Diabetic Retinopathy, Glaucoma, Retinal Imaging, Image Classification, Medical AI, Self-Supervised Learning, Machine Learning, Benchmarking, Vision Transformer.

# Contents

# Abstract

**Background**: Integrating deep learning into medical imaging is poised to greatly advance diagnostic methods, but it faces challenges with generalizability, which can foster distrust in artificial intelligence and exacerbate ethnic biases. Foundation models, based on self-supervised learning, address these issues, improve data efficiency and reduce ethnic biases. Natural domain foundation models show promise for medical imaging, but systematic research evaluating domain adaptation of these models, especially using self-supervised learning and parameter-efficient fine-tuning, remains underexplored. Additionally, little research addresses the issue of catastrophic forgetting observed during fine-tuning of foundation models.

**Methods**: We adapted the DINOv2 vision transformer for retinal imaging classification tasks using self-supervised learning and generated two novel foundation models termed DINORET and BE DINORET. Publicly available color fundus photographs from multiple datasets were employed for model development and subsequent fine-tuning for diabetic retinopathy staging and glaucoma detection. We introduced block expansion as a novel method for domain adaptation in retinal imaging and assessed the models for catastrophic forgetting. Models were benchmarked to RETFound, a state-of-the-art foundation model in ophthalmology.

**Findings**: DINORET and BE DINORET demonstrated competitive performance on retinal imaging tasks, with the block expanded model achieving the highest scores on most datasets. Block expansion successfully mitigated catastrophic forgetting. Our few-shot learning studies indicated that DINORET and BE DINORET outperform RETFound in terms of data-efficiency.

**Interpretation**: The study highlights the potential of adapting natural domain vision models to retinal imaging using self-supervised learning and block expansion. BE DINORET offers robust performance and high data efficiency without sacrificing previously acquired capabilities. Our findings suggest that these methods could enable healthcare institutions to develop tailored vision models for their patient populations, enhancing global healthcare inclusivity.

**Funding:** This research was supported by grants from the Spross Research Institute of the Werner H. Spross Stiftung zur Förderung der Augenheilkunde, Zurich, Switzerland, and the National Institute on Aging (1R41AG073066-01). SKW is funded by the Medical Research Council (MR/T000953/1). The study was further supported by the Center Core Grant P30EY01480 and Research to Prevent Blindness- Unrestricted Grant (GR004596-1).

# 1. Introduction

Integrating deep learning (DL) into medical imaging offers significant advancements in diagnostics (1–3). However, DL models face challenges with generalizability, particularly when facing distribution shifts between training datasets and clinical settings, leading to reduced trust in artificial intelligence (AI) among physicians and the public and substantial ethnic bias in DL models (4–11). Additionally, supervised training of DL models requires large, labeled datasets, which is often impractical and limiting (2,12,13). Self-supervised learning (SSL) addresses this by utilizing large datasets without annotation, thus reducing manual workload and expanding model development (14–17).

Foundation models (FMs), pre-trained on large image datasets using SSL, show robust feature representation capabilities and are easily adapted to various tasks (10,16,18–24). They are particularly beneficial for institutions with limited data and help mitigate ethnic bias, thus ensuring equitable diagnostic outcomes (6,10,16,24–28). Domain adaptation involves fine-tuning a pre-trained FM on a target domain with minimal additional training data (29–31). This process leverages the model's pre-existing knowledge while adjusting it to the nuances of the new domain. In the context of medical imaging, domain adaptation is crucial for ensuring that FMs can accurately interpret and analyze medical images, which often differ significantly from the data used during initial training (10,20,23,25).

Adapting natural domain FMs like the DINOv2 Vision Transformer (ViT) model to medical imaging can substantially improve DL models and data efficiency (32–35). Still, effective fine-tuning strategies are warranted to address domain shifts (32,34). However, while there is a growing interest and active research in adapting DINOv2 for various medical applications (19,35–38), no study has investigated SSL for natural domain adaptation for retinal imaging classification tasks.

Color fundus photographs (CFPs) are a widely utilized imaging modality in ophthalmology. They enable the diagnosis and monitoring of retinal diseases and offer insights into systemic health, such as the ability to predict major cardiovascular event (39–46). Efficient adaptation using SSL would enable the development of numerous FMs tailored to the needs of individual healthcare institutions with only moderate computational and image requirements (10).

Catastrophic forgetting, where models lose performance on original data after fine-tuning on new data, remains a challenge in machine learning (47–49). Adapting foundation models for retinal imaging while mitigating catastrophic forgetting is a crucial area of research. Parameter-efficient fine-tuning (PEFT) methods, like block expansion (BE), minimize trainable parameters while preserving model features, thereby enhancing generalizability and preventing catastrophic forgetting (50–52). To the best of our knowledge, we were the first researchers to explore BE in the context of PEFT in computer vision (53). Despite its success in language models, BE still has limited exploration in computer vision and medical AI (51–54).

Given DINOv2's promising results in medical image classification, we aim to explore SSL strategies to adapt it to CFPs and generate novel foundation models for Ophthalmology termed DINORET and BE DINORET. We propose methods for generating medical FMs using SSL and BE and compare these against RETFound (10), an influential FM in ophthalmology. This work makes the following contributions in five key areas:

1. SSL Strategies: Introduces SSL strategies for adapting natural domain ViTs to the medical domain and proposes methods for generating medical FMs using SSL and BE.
2. Model Architecture Practicality: Retains small model architectures to reduce computational demands, making the models practical and deployable in clinical environments.
3. Data Efficiency: Shows that DINOv2, DINORET and BE DINORET exhibit superior data efficiency over RETFound.
4. Performance: Demonstrates that our models consistently outperformed RETFound in all experiments with a frozen backbone. At the same time, unfrozen BE DINORET exceled and most frequently ranked among the top models. Additionally, BE preserved prior performance and feature representation capabilities, successfully avoiding catastrophic forgetting. The BE strategy allowed for continuous model improvement while retaining maximum generalizability.
5. Future Benchmarking: Suggests that future benchmarking of foundation ViT models in medicine should focus on the quality of embeddings rather than fine-tuning strategies to ensure fair comparisons and minimize overfitting.

# 2. Methods

## 2.1 Datasets

For the scope of this study, only publicly available CFP datasets were employed. All datasets are either freely available or can be obtained after registration. Ground truth labels for diabetic retinopathy (DR) stages (according to the International Clinical Diabetic Retinopathy Scale (ICDR)) (55) and the presence of glaucoma are all publicly available.

As shown in Table 1, we employed three datasets for SSL post-pretraining: Kaggle Eye Picture Archive Communication System (EYEPACS) (11,56,57), DDR (58), and Artificial Intelligence for RObust Glaucoma Screening (AIROGS) (59). The EYEPACS and DDR datasets contain images of healthy eyes and eyes with various stages of DR, while the AIROGS dataset includes images of healthy and glaucomatous eyes. We used the five datasets listed in Table 2 for downstream classification tasks. These include the Asia Pacific Tele-Ophthalmology Society (APTOS) (60), Methods to Evaluate Segmentation and Indexing Techniques in the Field of Retinal Ophthalmology (MESSIDOR-2) (61–63), Indian Diabetic Retinopathy Image Dataset (IDRiD) (64,65), The Diabetic Retinopathy Two-field image Dataset (DRTiD) (66) for DR and PAPILA (67) for glaucoma. More detailed information on each dataset can be found in Supplementary Material 1.

| Dataset Name | Total Number of Available Images | Total Number of Importable Images | Number of Images with Sufficient Quality (AutoMorph) | Data Type | Country of Origin |
|---|---|---|---|---|---|
| Kaggle-EyePacs | 88,702 | 88,699 | 70,734 | DR, Healthy | USA |
| DDR | 13,673 | 13,604 | 9,299 | DR, Healthy | China |
| AIROGS | 101,442 | 101,267 | 101,267 | Glaucoma, Healthy | USA |
| Total CFP Images (CFP-Large) | 203,817 | 203,570 | 156,074 | DR, Glaucoma, Healthy | USA, China |

**Table 1: CFP datasets employed for SSL post-pretraining.** This table lists the CFP datasets that were used during SSL for domain adaptation of ViTs in this study. Three datasets were utilized, with their names being listed in the first column. The table additionally lists the total number of CFPs within a dataset, the number of importable images after excluding damaged images, the number of images that were deemed to be of sufficient image quality by AutoMorph, the type of data included in the dataset (healthy or with eye pathology), and the country of origin for the images. Total CFP images (CFP-Large) in the last row refers to the compiled dataset for SSL in this study, obtained when pooling all images from the three datasets listed (DR=Diabetic Retinopathy).

| Dataset | Labels | Task | Total Number | Defined Test Set (n) | Split | Ungradable Image Count (%) | Country of Origin |
|---|---|---|---|---|---|---|---|
| APTOS | H, DR1-4 | DR | 3,662 | NO | 70:15:15 | 724, 19.77% | India |
| DRTiD | H, DR1-4, DME | DR | 1,550 | Yes (550) | 54:11:35 | 508, 33.66% | China |
| IDRiD | H, DR1-4, DME | DR | 513 | Yes (103) | 66:14:20 | 57, 11% | India |
| Messidor-2 | H, DR1-4, DME, Q | DR | 1,748 | NO | 70:15:15 | 52, 2.97% | France |
| PAPILA | G, S, H | Glaucoma | 488 | NO | 70:15:15 | NA | Spain |

**Table 2: Datasets used for supervised fine-tuning tasks.** This table compares the five CFP datasets used for downstream classification tasks in this study. The datasets include APTOS, DRTiD, IDRiD, Messidor-2, and PAPILA, with ground-truth labels ranging from Healthy (H) to different stages of Diabetic Retinopathy (DR1-4), Diabetic Macular Edema (DME), Suspected-Glaucoma (S), Glaucoma (G) and image Quality (Q). Each dataset's total number of images, the presence of a defined test set, the split ratios for training, validation, and testing, and the percentage and count of ungradable images according to AutoMorph are specified, and the country of origin for each dataset is listed.

## 2.2 Model Architectures

In this study, several ViT models (68) were compared, focusing on three fundamentally different architectures, as detailed in Table 3.

### RETFound

RETFound is a ViT with 24 transformer blocks, serving as the encoder of a pre-trained (ImageNet-16k (69)) masked-autoencoder (MAE). It is pre-trained on approximately 900,000 CFPs using a generative SSL approach (10). Parameters and model weights are publicly available, and the fine-tuning pipeline provided by the authors on GitHub was followed (70).

### DINOv2 ViT-B and DINORET

We utilized the base variant of the DINOv2 model, specifically the Vision Transformer Base (ViT-B), which comprises 12 transformer blocks with approximately 86.5 million parameters (33). DINOv2 is a ViT pre-trained on more than 146 million natural images using a contrastive form of SSL, as shown in Figure 1. DINORET has an identical architecture to DINOv2 but underwent additional training, as shown in Figure 1.

### Block-Expanded DINOv2 ViT-B (BE DINORET)

We used a novel method called block expansion (BE) for adapting pre-trained ViTs to the medical domain (50). As shown in Figure 1, BE duplicates existing transformer blocks and inserts them into the model (50). These blocks were initialized to act as identity operations, increasing the model's depth without changing its output before fine-tuning. During post-pretraining, duplicated blocks, each adding about 7 million parameters for DINOv2 ViT-B, were fine-tuned on new CFP images using a contrastive SSL approach, as detailed in section 2.4. This approach adapted only the duplicated blocks to the CFP domain during SSL, keeping the 12 original transformer blocks frozen (i.e., without allowing the backbone (BB) network to adapt during training) to retain pre-trained features (50,51).

| Base-Model | Number of Transformer Blocks | Total Number of Trainable Parameters (unfrozen BB) | Total Number of Trainable Parameters (frozen BB) | Post-Pretraining Data |
|---|---|---|---|---|
| RETFound | 24 | 303 M | 5,125 | CFP-Large, ImageNet-1k |
| DINOv2 ViT-B | 12 | 86 M | 3,845 | DINOv2: Natural Domain (LVD-142M) |
| DINORET ViT | | | | DINORET: Natural Domain (LVD-142M), CFP-Large |
| BE DINORET ViT | 15 | 107 M | 3,845 | 12 Blocks: Natural Domain (LVD-142M), 3 Blocks: CFP-Large |

**Table 3: Model Architectures and parameter counts.** This table depicts the parameter counts and number of transformer blocks for all major base models utilized in this study. The number of trainable parameters is also indicated, which differs between the two fine-tuning strategies used and is shown separately for models with frozen and unfrozen BBs. The number of trainable parameters when unfreezing the BB is identical to the total parameter count for each model. Several versions of each base model were evaluated, as described elsewhere in this manuscript.

**Figure 1: Overview of domain adaptation and architectures for DINORET, BE DINORET, and DINOv2.** This figure illustrates the architectural design of the DINOv2 ViT-B and methods for domain adaptation of DINORET and BE DINORET to CFPs during post-pretraining. (A) The DINOv2 ViT-B architecture comprises 12 transformer blocks with multi-head self-attention mechanisms and multilayer perceptrons (MLPs), including approximately 86.5 million parameters. The DINOv2 model is pre-trained on 146 million natural domain images using contrastive SSL and produces a single classification [CLS] token and multiple patch embeddings from each input image. (B) DINORET is a fully CFP-adapted model derived from DINOv2 ViT-B, where all weights from the pre-training phase on natural domain images are initialized and updated during SSL post-pretraining using our modified DINOv2 pipeline. (C) BE DINORET expands the DINOv2-ViT-B model to the CFP domain by duplicating existing transformer blocks and inserting them into the model. Before SSL post-pretraining, attention weights within duplicated blocks are preserved, and the linear projection layers are zero-initialized (indicated in pink), keeping the model's output unchanged. During SSL post-pretraining, all parameters within expanded blocks are updated on CFPs with our modified DINOv2 method (white), while the original blocks remain frozen (cyan). Each expanded transformer block adds approximately 7 million parameters. BE DINORET expands the original ViT to the CFP domain while aiming to preserve knowledge acquired in previous domains to avoid catastrophic forgetting (53). Trainable parameters during SSL post-pretraining are indicated in white, frozen in cyan, and zero-initialized, but trainable parameters in pink. GELU refers to a Gaussian Error Linear Unit activation function applied within the architectures. When used as input for SSL post-pretraining, CFPs were split into patches. The figure is adapted from Bafghi et al. (51), and permission was granted.

## 2.3 Pre-Processing

The three datasets we used for SSL post-pretraining were aggregated into a single CFP-Large dataset. With one exception (used for the ablation study), low-quality images were removed from the dataset using part of the AutoMorph pipeline (71,72). In addition, AutoMorph was used to crop the images to fit the round shape of CFPs, as shown in Supplementary Figure 1. Every CFP image was split into a variable number of patches. During the supervised fine-tuning, the only preprocessing operation was resizing the images to 224x224 pixels.

## 2.4 Self-Supervised Post-Pretraining Method

In our study, post-pretraining refers to updating the weights in the BB of the natural domain pre-trained ViT DINOv2 with SSL. The same post-pretraining method (contrastive SSL with a modified DINOv2 pipeline) was used for the DINORET and BE DINORET models, with one key difference regarding the architecture (33,73). For BE DINORET, BE was included, and all the weights of the original transformer blocks were frozen, while only duplicated blocks were updated. For DINORET, all weights were unfrozen. DINOv2 did not undergo any additional SSL before supervised fine-tuning. A detailed explanation of SSL post-pretraining approaches is provided in Supplementary Material 2.

## 2.5 Supervised Fine-Tuning

Two fundamentally different methods were used during task-specific fine-tuning (e.g., DR staging) with supervised learning (SL) on the target datasets.

**Frozen BB Fine-Tuning:** Frozen BB fine-tuning, also known as linear probing, involved keeping the weights of the BB layers fixed and only updating the parameters of an additional linear classification head, as shown in Figure 2 (13,23,34,51,74).

**Unfrozen BB Fine-Tuning:** Unfrozen BB fine-tuning involved updating the parameters of the entire model, including the BB layers, during supervised training, as illustrated in Figure 2 (13,32,34,51). The gradients derived from the task-specific loss function are propagated through all model layers, enabling parameter updates and fine-tuning the entire network to the new data (51).

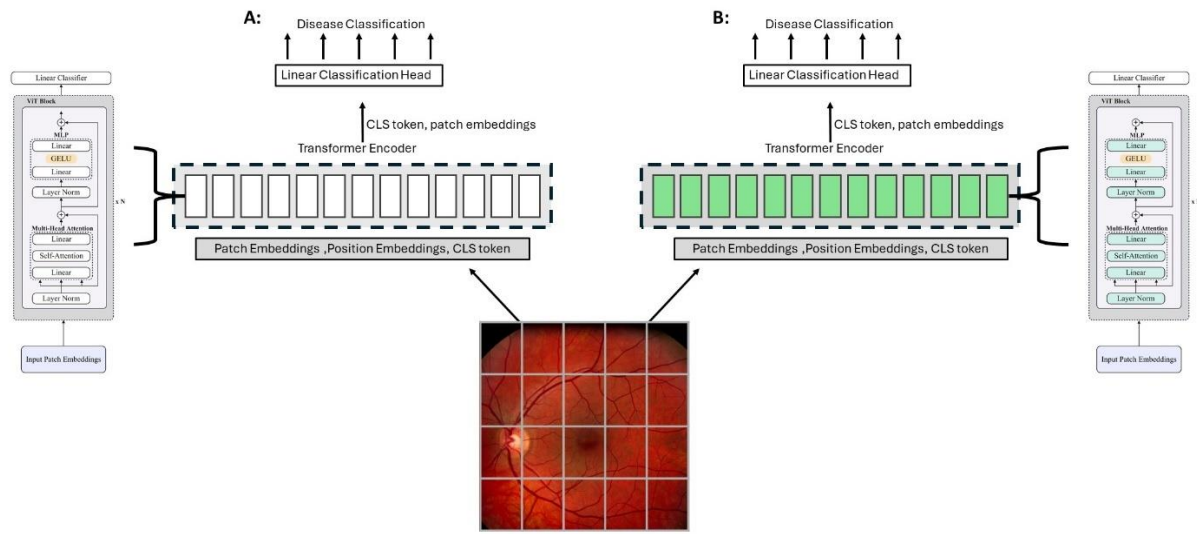

**Figure 2: Supervised Fine-Tuning for the downstream tasks of DR staging and glaucoma detection.** During task-specific model training with the CFPs (all containing ground-truth labels) are split into patches, each embedded with positional information, and a [CLS] token is appended. The parameters within the linear classification head are constantly updated during supervised fine-tuning for a specific task (shown in white). This head maps the high-dimensional output of the ViT (specifically the [CLS] token or patch embeddings) onto the desired number of disease classes. (A) The frozen BB fine-tuning approach (linear probing) is depicted. Only the parameters within the linear classification head are updated during task-specific fine-tuning, while weights within all transformer blocks of the BB encoder are frozen (shown in cyan). This method preserves the embeddings produced by the encoder BB during task-specific training, treating the BB as a fixed feature encoder. (B) Unfreezing BB layers during supervised training is depicted. All parameters, including weights in all layers of the BB (shown in white) and the linear classification head, are updated during supervised training. The gradients derived from the task-specific loss function are propagated through the entire model, optimizing all parameters in the BB and the resulting embeddings for the task at hand. Trainable parameters are depicted in white, while frozen parameters are shown in cyan. This approach (including color labels and transformer architecture) applies analogously to BE DINORET, which includes three additional transformer blocks in the BB encoder.

## 2.6 Image Classification

Disease classification was performed using a linear classification head consisting of a linear layer followed by a softmax function. This layer projects the high-dimensional output of the ViT, such as the [CLS] token or average patch embedding, to the desired number of target classes.

## 2.7 Hyperparameters

The exact supervised fine-tuning procedure was similar among all models, except for RETFound, where we used the hyperparameters suggested by Zhou et al. (10,70). The hyperparameters used for training the DINOv2-based models are detailed in Supplementary Material 3, and the final models were always chosen based on the best qKappa checkpoint on the validation set.

## 2.9 Evaluation Metrics and Hypothesis Testing

We reported metrics that ensure optimal comparison with other publications, including the F1 score, overall accuracy (Acc), Area Under the Receiver Operating Characteristic Curve (AUC ROC), and quadratic weighted Kappa (qKappa) (75). As commonly used in DR classification (11,62), a binary classification score was introduced, where no DR (stage 0) and mild DR (stage 1) were collectively grouped as non-referable DR. In contrast, moderate (stage 2), severe (stage 3), and proliferative DR (stage 4) were grouped as referable DR (rDR) (62,76,77). To evaluate the performance on the original pre-training domain, the k-nearest neighbors (kNN) score on the validation set of the ImageNet-1k dataset was reported (69,78). The formulas used to obtain these metrics can be found in our source code or Supplementary Material 3. We refrained from inferential statistics for experiments lacking repeated runs and having invariant splits. For experiments with variable seeds and multiple independent runs, hypothesis testing was conducted using either ANOVA or Kruskal-Wallis tests based on the data's distributional characteristics, assessed by the Shapiro-Wilk (for normality) and Levene tests (for homoscedasticity). Post-hoc comparisons were performed using Tukey's Honestly Significant Difference (HSD) test following ANOVA, which adjusts p-values for multiple comparisons, and Dunn's test following Kruskal-Wallis, with the latter applying Benjamini-Hochberg (BH) corrections for multiple comparisons and both raw and adjusted p-values were reported. Additionally, hierarchical linear mixed models (LMMs), estimated via maximum likelihood (ML), were fit across pooled data, and a likelihood ratio test (LRT) was used to compare the full models against reduced models. The significance level for all tests was set at an alpha of 0.05.

## 2.10 Experiments

### 2.10.1 Hyperparameter Studies

**SSL Parameters:**
We assessed the impact of filtering the CFP-Large dataset using AutoMorph and excluding bad-quality images as defined by the algorithm (71). We also analyzed the effect of expanding BE DINORET by a varying number of blocks (specifically, 1-12 blocks) and evaluated models trained with smaller or larger patches (1-25 patches). Models were tested on the full APTOS dataset and a few-shot study with 16 sample images per class was performed.

**Supervised Fine-tuning Parameters:**
We tested the influence of using either the [CLS] token, the average of all patch embeddings, or both concatenated for image classification. Additionally, we investigated the use of a distance-weighted cross-entropy loss (CEL) compared to a vanilla CEL for classification tasks.

### 2.10.2 Model Evaluation Studies

**DR Classification Tasks**

- **Multi-Source Domain Fine-Tuning (MSDFT)**: Following Met et al. (77), we combine all DR datasets into unified training, validation, and test sets. Models are fine-tuned on the joint training set and evaluated on individual and joint test sets.

- **Performance Assessment on Test Sets**: Each model is evaluated on its test set to establish baseline performance.
- **Cross-Evaluation Performance**: Models are trained on one DR dataset and tested on all others to assess robustness to population shifts and variations in image acquisition.

**Glaucoma Detection Task**

- **Performance Assessment on Test Set**: Each model is evaluated on the PAPILA dataset to determine baseline performance.

**Data Efficiency through Few-Shot Learning**

- **Few-Shot Learning Study**: We conduct a few-shot learning study on the four DR datasets and the glaucoma dataset by reducing the training subset to a fixed number of samples per class while keeping validation and test sets unchanged. All runs are performed in quintuplicate, with random selection of training images.

## 2.10.3 Catastrophic Forgetting

To determine if our SSL-based strategies for domain adaptation of DINOv2 to CFPs make models susceptible to catastrophic forgetting, DINORET, BE DINORET, and DINOv2 were tested on the validation subset of the ImageNet-1k dataset using the kNN algorithm with k=20 of the embeddings ([CLS]) token. To investigate if the embeddings generated from CFPs by DINORET and BE DINORET substantially differ from unmodified DINOv2, we retrieved all embeddings ([CLS] tokens) generated by the three models on the test dataset of APTOS and performed a linear discriminant analysis (LDA) of the embeddings across all dimensions.

# 3. Results:

## 3.1 Hyperparameter Studies

AutoMorph classified 47,496 images in the CFP-Large dataset as ungradable. Removing these ungradable images improved DR staging performance on APTOS across all metrics except for AUC ROC, which remained unchanged (Supplementary Data 1). Supplementary Figure 2 shows that varying the number of transformer blocks in BE DINORET led to inconsistent evaluation results while splitting CFPs into patches during post-pretraining enhanced performance (Supplementary Figure 3). Supplementary Data 1 and Supplementary Results 1 provide comprehensive results from the studies on SSL and SL hyperparameters.

## 3.2 Multi-Source Domain Fine-Tuning (MSDFT)

MSDFT was performed exclusively for DR staging using a combined training and validation set of 4724 images from all four DR datasets. MSDFT fine-tuned models were evaluated on all individual test sets. All results obtained from our MSDFT experiments are depicted in Figure 3. The confusion matrices can be found in Supplementary Figure 4. Unfrozen models consistently outperformed their frozen counterparts for qKappa and in 95% of experiments for AUC ROC (85% for Acc, 90% for F1, 95% for rDR Acc).

Overall, BE DINORET with an unfrozen BB showed the best results with the highest qKappa and AUC ROC scores on the Messidor and IDRiD datasets and, additionally, the best qKappa score on the combined testing subset, as shown in Figure 3. As depicted in Figure 4, unfrozen BE DINORET achieved the highest percentage of being the best-performing model on all five datasets for all metrics, except for AUC ROC (where DINOv2 excelled), Acc and F1 (where DINORET or RETFound, respectively, were equally often the best-performing models).

Unfrozen DINORET displayed the best qKappa score on DRTiD, while unmodified DINOv2 had the best qKappa and AUC ROC score on APTOS and the best AUC ROC score on the pooled subset. RETFound did not achieve a leading qKappa or AUC ROC score on the testing datasets.

Regarding rDR Acc, BE DINORET had the best score on all testing datasets, except for APTOS, where DINOv2 performed better.

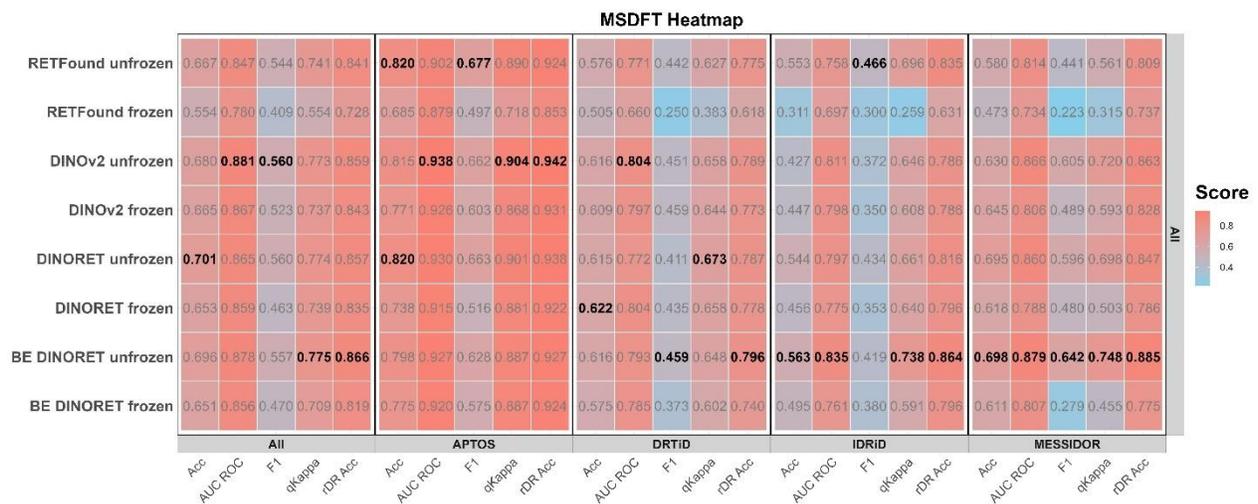

**Figure 3: MSDFT results by model.** A heatmap depicting the results from the MSDFT experiments, where models were trained on the pooled training subsets of the 4 DR datasets and evaluated on the testing subsets of each dataset individually or pooled (All). Datasets on which the MSDFT fine-tuned models were tested are segregated along the x-axis and annotated in the gray bars. Model types and the evaluation metrics are listed on the y-axis and x-axis, respectively. The color gradient represents the score for each metric, ranging from light blue tones (lower scores) to red (higher scores), with the best model per metric (column) being indicated in bold.

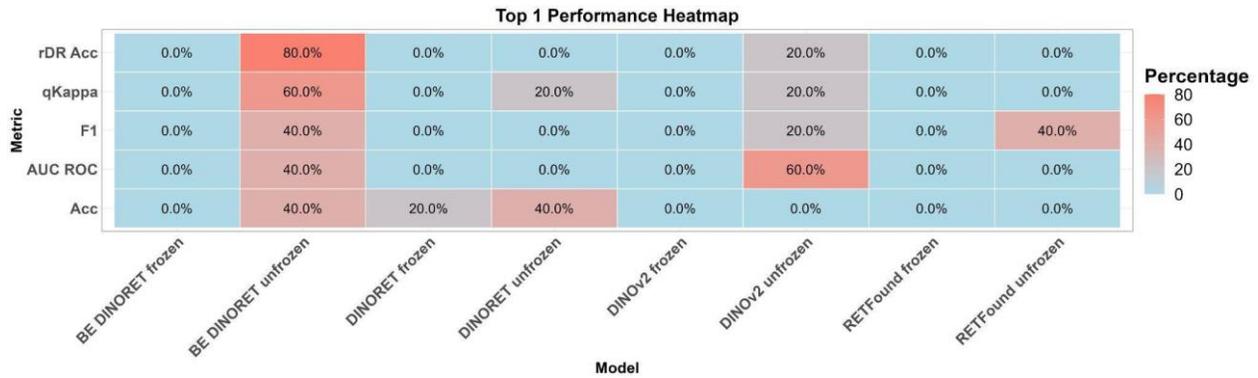

**Figure 4: Top 1 performance heatmap.** Unfrozen BE DINORET most commonly ranks as the best MSDFT fine-tuned model. A heatmap visualizing the percentage of test DR datasets (out of 5) that an MSDFT fine-tuned model ranked first on for each metric. Model types and the evaluation metrics used are listed on the x-axis and y-axis, respectively. The color gradient represents the percentage of datasets a single model ranks first on, ranging from light blue (lower) to red tones (higher). The exact percentage values are displayed on the tiles.

## 3.3 Performance Assessment on Test Sets

In this assessment, supervised training, validating, and testing on a single dataset was performed for 5 datasets (4 for DR and 1 for glaucoma). Unfrozen models outperformed their frozen counterparts 80% of the time for AUC ROC, 90% for qKappa, Acc, and F1, and in all cases for rDR Acc. For unfrozen models, results varied by metric and dataset, as displayed in Figure 5. BE DINORET displayed the best qKappa score on Messidor and the best AUC ROC score on PAPILA. DINORET had the best AUC ROC score on Messidor and DRTiD, while DINOv2 had the best qKappa score on IDRiD and DRTiD. RETFound achieved the best AUC ROC and qKappa scores on APTOS and the best AUC ROC score on IDRiD. As illustrated in Figure 6, unfrozen BE DINORET most commonly ranked amongst the two best performing models, only being matched by RETFound for AUC ROC and Acc and by DINOv2 for F1.

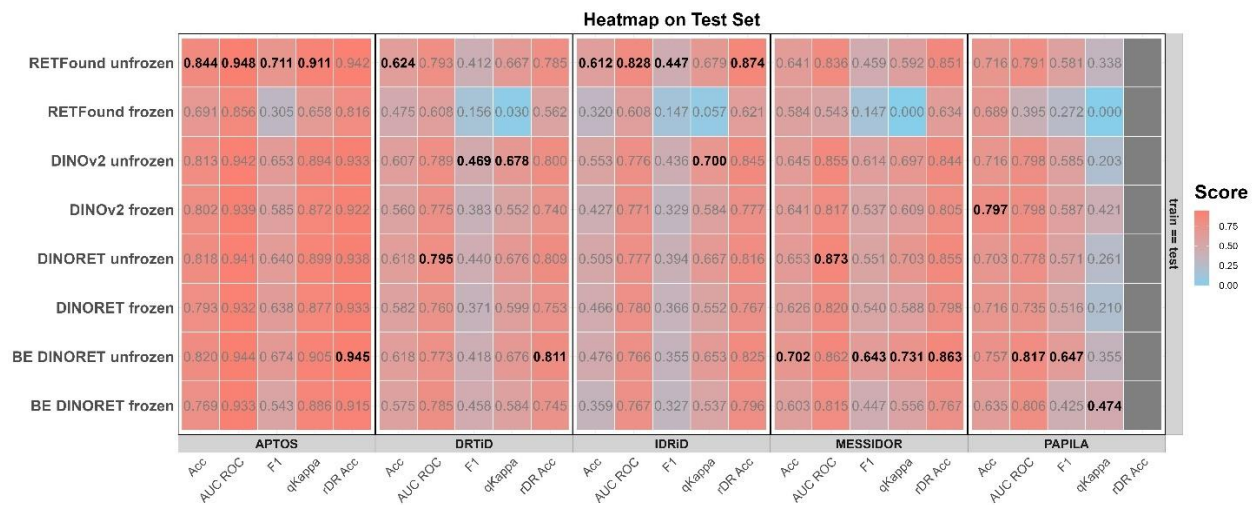

**Figure 5: Performance assessment on test sets by model.** The heatmap illustrates the performance of all models on the 5 distinct datasets (4 for DR, 1 for glaucoma), segregated along the x-axis and annotated in the gray bars. The x-

axis additionally lists the evaluation metrics, while the y-axis lists the models and their backbone states (frozen or unfrozen). The color gradient represents the score values, with higher scores in red and lower in light blue tones. The highest score per metric (column) is highlighted in bold.

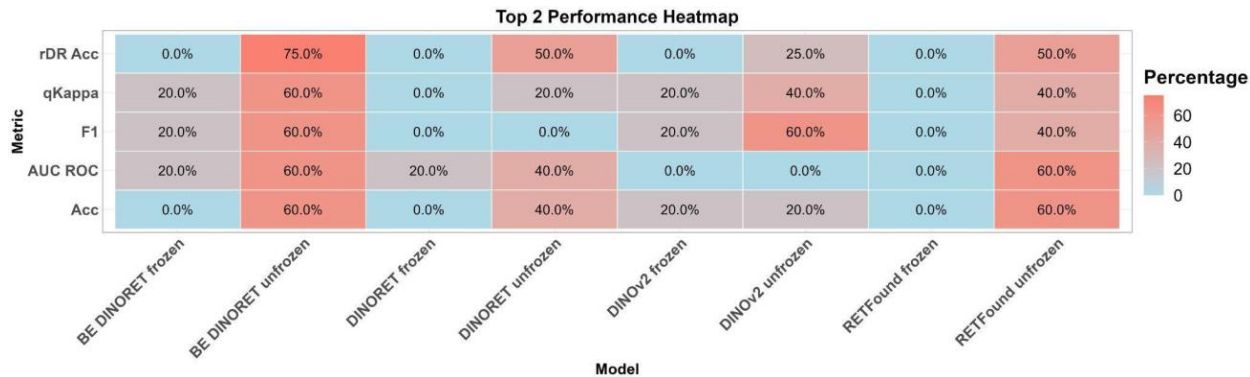

**Figure 6: Top two performance by model and evaluation metric for performance on test dataset.** Unfrozen BE DINORET most commonly ranks amongst the best two models when training and testing on a single dataset. The heatmap visualizes the percentage of datasets (n=5) that each model ranked among the two best models (Top 2) on for a specific evaluation metric. The x-axis lists the models, while the y-axis lists the metrics. The color gradient represents the percentage of the Top 2 rankings by model, with higher percentages in red tones and lower in light blue. Text labels within the cells provide the exact percentage values.

## 3.4 Cross-evaluation:

Compared to previous experiments in this study, frozen models performed better during cross-evaluations, outperforming their unfrozen counterparts in 39.6% for AUC ROC and 20.84% for qKappa (45.8 % for Acc, 43.75% for F1, 29.2% for rDR Acc). As shown in Supplementary Figure 5, when averaging scores on a single test dataset across the three cross-evaluations, RETFound most frequently achieved the highest average (avg) AUC ROC and avg qKappa scores, followed by the other models. A detailed depiction of all results can be found in Supplementary Figure 6 and Supplementary Results 2.

## 3.5 Data Efficiency

All runs were performed in quintuplicate until the sample count could not be increased further, resulting in up to 16 training images per class for Messidor, IDRiD, and DRTiD and up to 32 or 128 images for PAPILA and APTOS, respectively. Supplementary Figures 7-11 depict the average score (across 5 runs) achieved by the models at a given training sample count. Supplementary Data 3 includes all results obtained from these experiments. Our models' precision across the 5 replicates significantly increased with training sample counts, alongside a statistically significant increase in qKappa and AUC ROC and a significant increase in performance when unfreezing BBs (all p = < 0.01, LMM with an LRT).

### Unfrozen Backbones:

Overall, for unfrozen fine-tuning between sample sizes of 4-64, BE DINORET most frequently ranked as the model with the highest AUC ROC score, followed by DINORET and DINOv2, and lastly, RETFound (Figure 7). Regarding qKappa, DINORET outperformed other models, most frequently achieving the highest

qKappa score, succeeded by DINOv2, RETFound, and BE DINORET, as illustrated in Figure 7. Supplementary Data 4 depicts the raw data from these experiments.

BE DINORET and DINORET consistently outperformed RETFound on all DR datasets for both AUC ROC and qKappa for image counts between 2 and 16, except for n=16 on IDRiD and Messidor, as shown in Figure 8. However, differences were only significant for some comparisons, as shown in Supplementary Data 5 (ANOVA or Kruskal-Wallis with Tukey HSD or Dunn post-hoc comparisons). At image counts beyond 16, the BE DINORET and DINORET models outperformed RETfound at n=32 and n=64 for AUC ROC, and DINORET outperformed RETFound at n=124 for qKappa, with all differences lacking statistical significance (ANOVA or Kruskal-Wallis test). Regarding glaucoma detection, except for n=1, BE DINORET and DINORET outperformed RETFound at all sample counts, though insignificant (ANOVA or Kruskal-Wallis). Additional analysis with the respective statistical results can be found in Supplementary Results 3.

## Frozen Backbones:

On all datasets (for both glaucoma detection and DR staging), BE DINORET, DINORET, and unmodified DINOv2 significantly outperformed RETFound for all metrics, as shown in Supplementary Figures 7-11 and Supplementary Data 5 (ANOVA or Kruskal Wallis with Tukey HSD or Dunn post-hoc comparisons). However, differences across BE DINORET, DINORET, and unmodified DINOv2 were inconsistent and all insignificant, as illustrated in Supplementary Data 5. As shown in Figure 7 (A) and (B), DINORET most frequently ranked as the model with the highest AUC ROC and qKappa scores, followed by BE DINORET and DINOv2. RETFound did not achieve a highest score for any task.

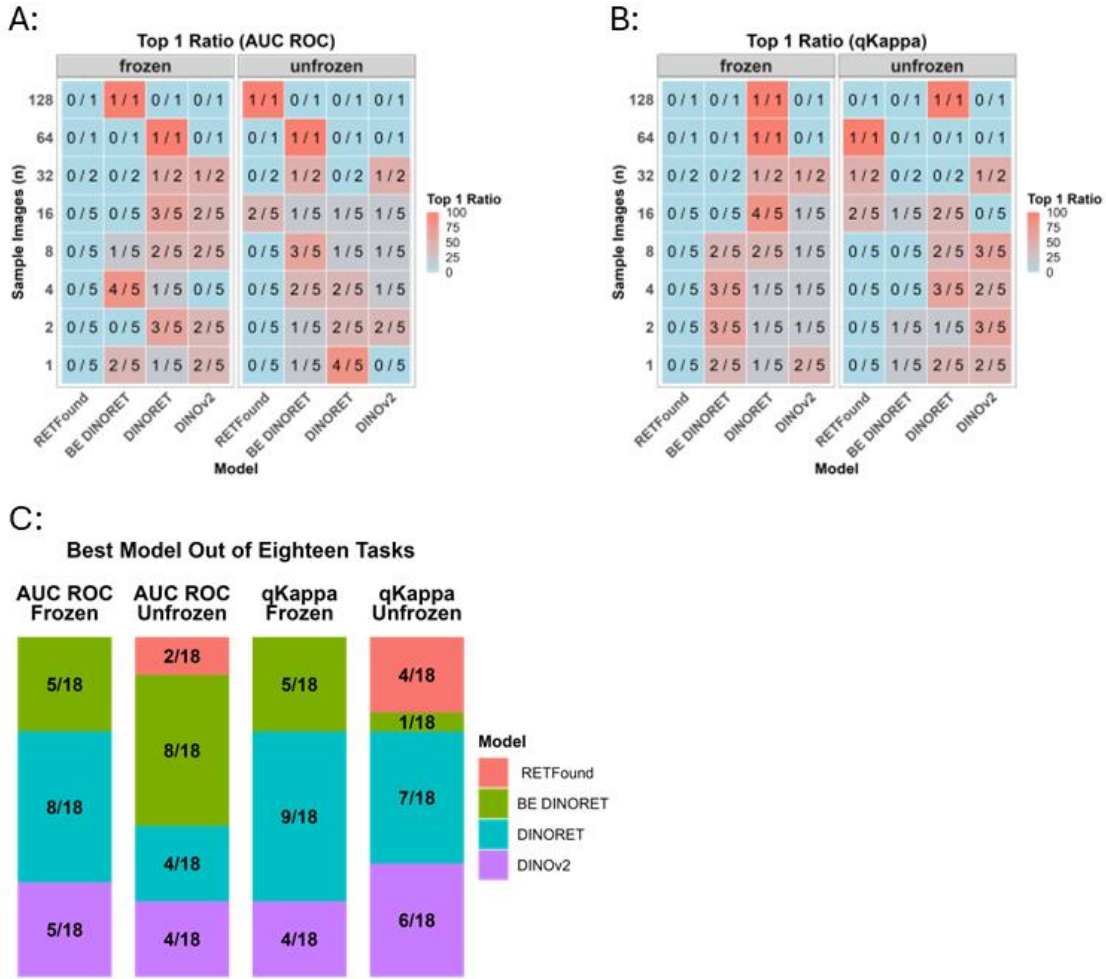

**Figure 7: Number and ratio of datasets a model ranks best on by evaluation metric and training sample count.** These figures present the results obtained from our few-shot experiments on 5 datasets (4 for DR, 1 for glaucoma) ranging from 1 to 128 training instances per class. Due to image scarcity, analysis beyond 16 instances per class was restricted to PAPILA and APTOS. All runs were performed in quintuplicate, and the scores obtained were averaged. Models were ranked within each backbone (BB) state across all datasets at a given sample count, and the model with the highest score for an evaluation metric was determined. Figures 1 A and B detail the distribution of the highest-ranking model and the total number of datasets for a given few-shot count, segregated by the BB state for AUC ROC (Figure 1 A) and qKappa (Figure 1 B). The color gradient represents these ratios, with higher values in red and lower in light blue. Figure 1 C aggregates the performances for few-shot counts between 4 and 64, where a total of 18 tasks were assessed. Here, the sum of instances where each model ranked highest was computed and expressed as a ratio of the total 18 possible instances. Models are color-coded, and the heights of the bars reflect the proportion of tasks for which each model attained the highest ranking. The visualization is segmented by evaluation metric (AUC ROC and qKappa) and BB state (Frozen and Unfrozen).

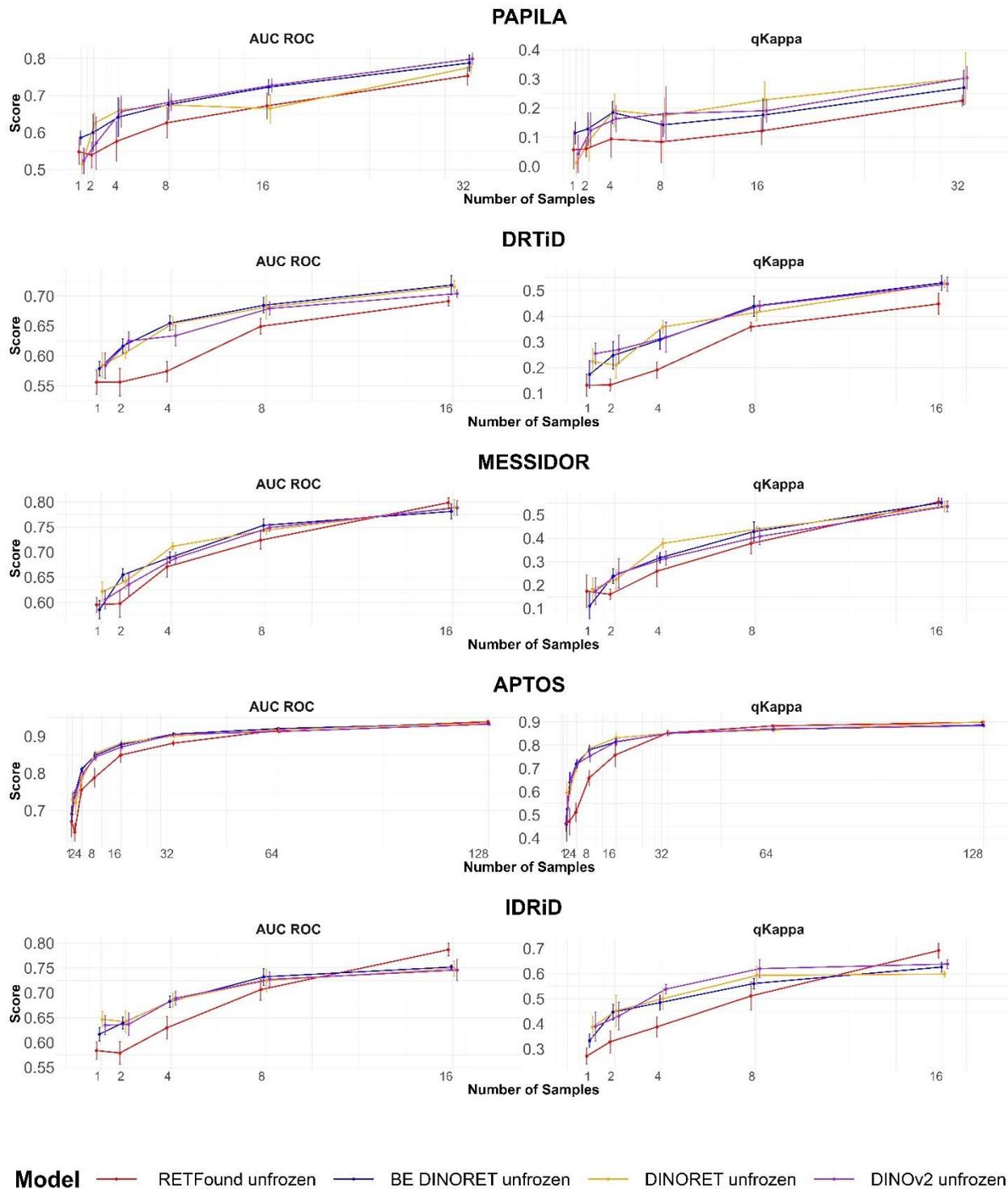

**Figure 8: Data efficiency for unfrozen fine-tuning.** The figures illustrate the results obtained from our few-shot experiments on all 5 datasets utilized when unfreezing the BB during supervised training. Each subplot represents a distinct dataset. The x-axis represents the number of sample images per class (few-shot), while the y-axis indicates

the performance score for the respective metric. Models are color-coded, and error bars denote the SEM across 5 runs. The plot titles correspond to the datasets used, and subtitles refer to the evaluation metrics (AUC ROC and qKappa).

## 3.6 Catastrophic Forgetting

While SSL domain adaptation of DINORET diminished its performance on the original domain, i.e. natural images, BE DINORET did not suffer from such a decrease and retained previously acquired capabilities, as shown in Table 4.

| Model | Top-1 Accuracy [%] | Top-5 Accuracy [%] |
|---|---|---|
| DINOv2-ViT-B (unmodified) | 93.916 | 82.110 |
| DINORET | 92.908 (-1.008) | 79.996 (-2.114) |
| BE DINORET | 93.966 (+0.050) | 82.070 (-0.040) |

**Table 4: ImageNet-1k accuracies for DINOv2, DINORET and BE DINORET.** The Table shows the kNN evaluation on the ImageNet-1k validation dataset using k=20. The original DINOv2 model is compared to the two domain-adapted versions (DINORET, BE DINORET), and both Top-1 and Top-5 Accuracy scores are listed for each model. Differences to the baseline performance of DINOv2 are indicated in brackets

As shown in Figure 9, the embeddings generated from the DR images of the APTOS test set by the DINOv2 ViT-B model appear separated, with distinct clusters for different DR stages. The DINORET model shows similar clustering, with slightly more overlap between some classes. BE DINORET generates embeddings that differ only slightly from the ones generated by unmodified DINOv2, but especially for images with higher DR stages, it demonstrates better-defined clusters with less overlap for different ground truth labels.

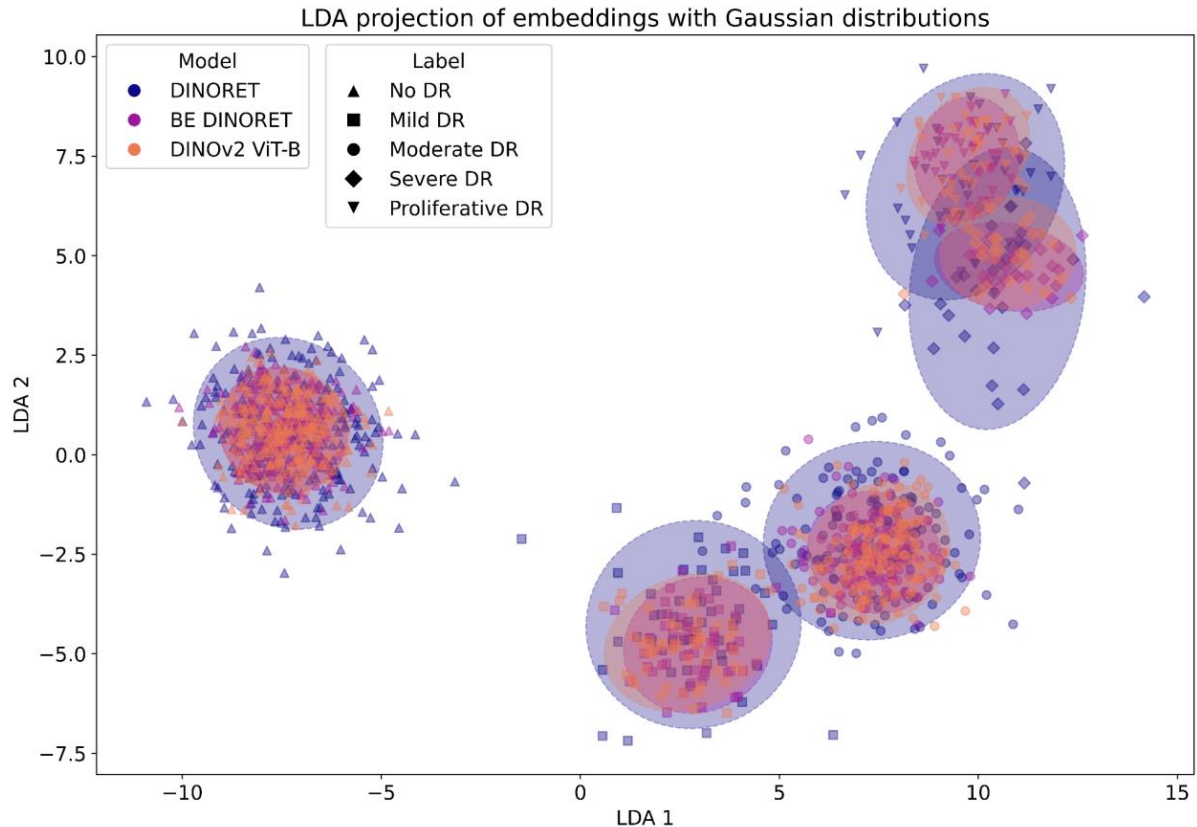

**Figure 9: Projected embeddings from the APTOS dataset by ViT models.** This figure shows the embeddings ([CLS] token) of each CFP from the test set of APTOS in a 2D subspace generated with a linear discriminant analysis (LDA). Each sample image is plotted 3 times using the distinct embeddings generated by each model (DINORET, BE DINORET, and DINOv2 ViT-B). Models are indicated by color, shapes correspond to ground truth labels of an image, and for each ground truth label (DR stage), an ellipse is overlaid that corresponds to twice the standard deviation (SD) from the mean of each class by model.

# 4. Discussion

Our research highlighted that DINOv2 exhibits remarkable out-of-the-box performance on retinal image classification tasks (32–34). When fine-tuning models with a frozen BB, DINOv2 outperformed RETFound in all performed experiments. Although unfrozen RETFound seemingly outperformed DINOv2 when training and testing on a single dataset or during cross-evaluations, in MSDFT experiments DINOv2 outperformed RETFound (10). Additionally, DINOv2 demonstrated superior data efficiency over RETFound.

We introduced strategies for adapting natural domain ViTs to the medical domain with SSL. Our findings demonstrated that these novel ViTs can improve performance for classification tasks, especially domain adaptation with BE and SSL, which displayed considerable promise. Unfrozen BE DINORET most commonly ranked as the best model for MSDFT experiments, and it most frequently was among the best two models when training and testing on a single dataset. When evaluating data efficiency, frozen BE DINORET outperformed frozen DINOv2, but these differences were not apparent when unfreezing the BBs. Lastly, BE DINORET achieved the highest qKappa and AUC ROC scores overall recorded on a dataset in this study for Messidor-2, IDRiD, and PAPILA. At the same time, no other model produced an overall best score on more than a single dataset. Our second model, DINORET, also exhibited characteristics distinct from DINOv2, potentially improving data efficiency and cross-evaluation performance.

Expanding models to novel domains with SSL while mitigating the risk of catastrophic forgetting is paramount, allowing healthcare institutions to develop DL models tailored to their patient demographics (22,25,51). Our findings indicated that simple retraining of all weights in DINOv2 on CFPs diminished performance on the previous domain. Conversely, BE preserved performance and avoided catastrophic forgetting. Thus, BE could potentially enable fine-tuning models on data from individuals of a specific age, gender, or ethnicity, without suffering a decrease in performance on data from individuals with a different demographic distribution. The ability of BE to maintain high generalizability and retain prior knowledge marks an essential step towards achieving a globally inclusive healthcare system and equitable access to diagnostic tools (20,21).

Maintaining a small model architecture, reduces computational demands and enhances the practicality of deploying these models in clinical environments (10,21,34,79). DINORET and BE DINORET significantly outperformed RETFound when the BB was frozen, allowing for fine-tuning with less than 6000 trainable parameters, thereby keeping VRAM requirements low (23). Even when unfreezing all layers, the number of trainable parameters remained a fraction of the ones in RETFound, making our models practical and widely deployable in a clinical setting (10).

Future benchmarking of foundation ViT models in medicine should focus on the quality of generated embeddings, rather than the optimization of fine-tuning strategies. This shift will facilitate fair comparisons by emphasizing the model's intrinsic capabilities instead of focusing on task-specific fine-tuning, thus minimizing the risk of overfitting and ensuring adequate benchmarking. For this reason, we opted for a single linear layer as our classification head, instead of multiple layers or an additional transformer. After all, we anticipate that the main goal for foundational ViTs in medicine is that the feature representation capabilities of ViTs and the resulting embeddings are robust enough for good performance, irrespective of supervised fine-tuning strategies.

# 5. Conclusion

We demonstrate that the natural domain FM DINOv2 performs strongly for retinal imaging classification tasks, outperforming a state-of-the-art domain-specific model in most experiments conducted in this study (10). Additionally, we proposed two new pipelines for domain adaptation of natural domain FMs to the clinical domain with SSL and show increased performance of our domain adapted models DINORET and BE DINORET. This approach will allow healthcare institutions across the globe to develop FMs customized to their patient population and help mitigate the prevalent biases in medical AI and our models mark an important addition to the field of AI in ophthalmology. Furthermore, we show that our proposed BE strategy avoids catastrophic forgetting and allows for continuous model improvement while retaining maximum generalizability. BE could potentially allow model optimization for specific patient populations without sacrificing knowledge of images from a different demographic, thereby ensuring fair access to medical AI across ethnic and socioeconomic groups. Our approach paves the way for adapting natural domain FMs to medical imaging with SSL; however, future work should leverage more data for model development and refinement of domain adaptation strategies.

# 6. Supplementary Figures:

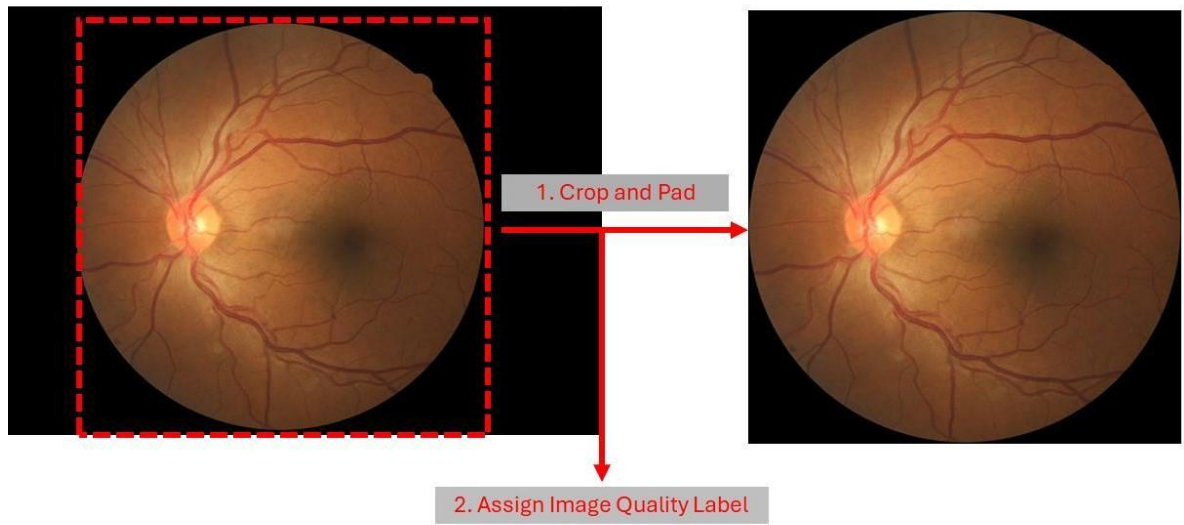

**Supplementary Figure 1: Pre-processing images from CFP-Large.** Images used for SSL during domain adaptation of our models were pre-processed with AutoMorph to crop them into a suitable shape, as depicted above. On the right-hand side, an image before pre-processing is depicted, while the image on the left-hand side corresponds to the processed image. Additionally, AutoMorph was used to predict the image quality grades of the CFPs.

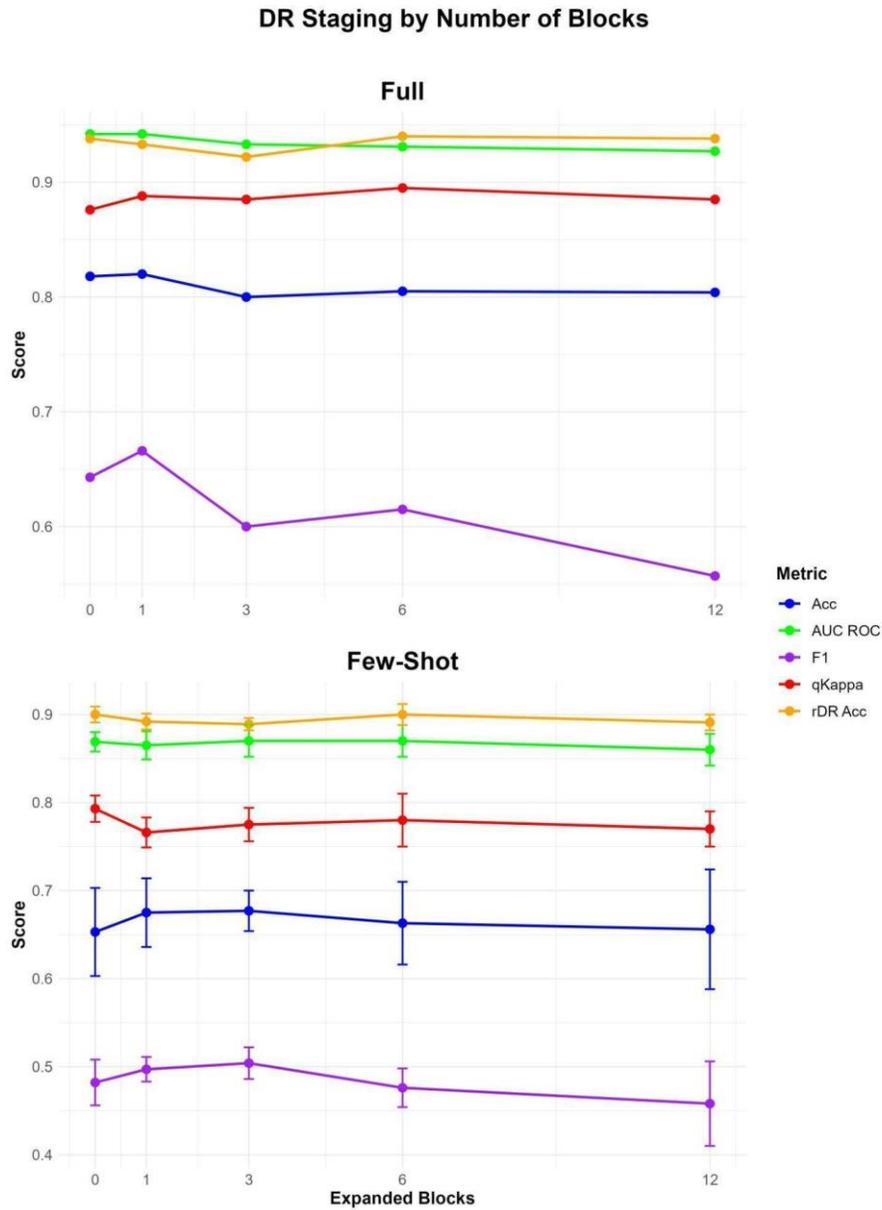

**Supplementary Figure 2: Number of expanded transformer blocks in BE DINORET.** This figure illustrates the effect of expanding an increasing number of transformer blocks in BE DINORET, ranging from 0 to 12 (0, 1, 3, 6, 12), on the performance of DR staging. Metrics evaluated include Accuracy (Acc), AUC ROC, qKappa, F1, and rDR Accuracy (rDR Acc). The top panel shows the results obtained after fine-tuning the BE DINORET models on the full APTOS dataset. The bottom panel presents the average of five runs from the few-shot study on the APTOS dataset with 16 training images per class and includes error bars representing SDs. Evaluation metrics are color-coded, as depicted in the graph legend.

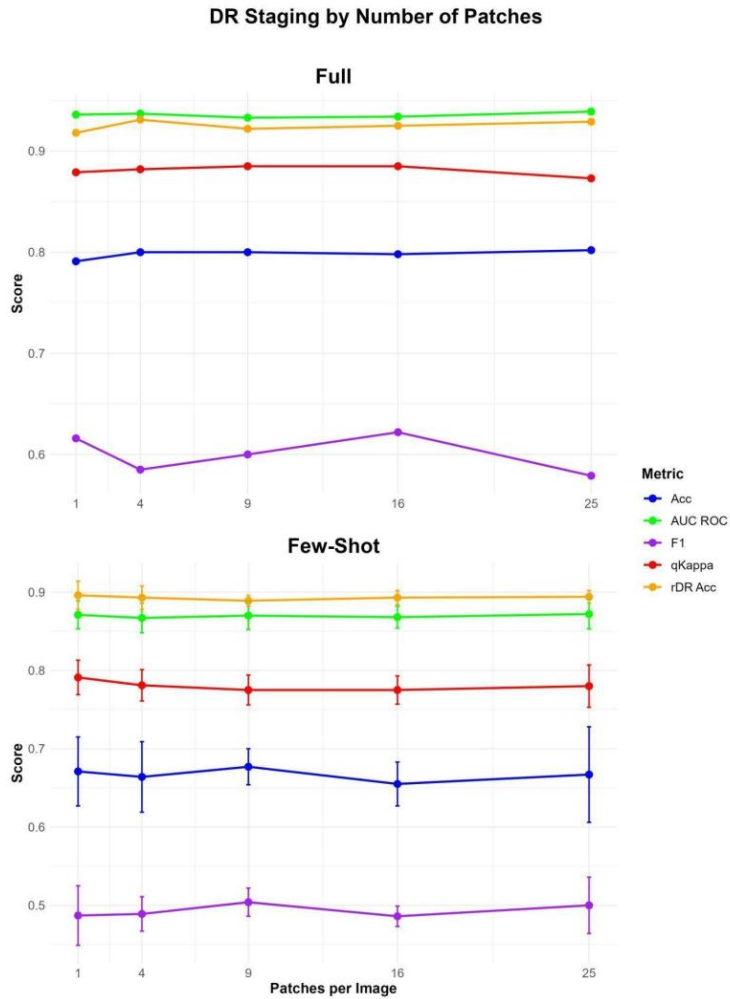

**Supplementary Figure 3: Splitting CFPs into patches during SSL improves subsequent DR staging on APTOS.** This figure illustrates the effect of splitting images into patches when post-pretraining BE DINORET on CFPs. CFPs were divided into 4, 9, 16, or 25 patches in grids ranging from 2x2 to 5x5. These approaches were subsequently evaluated by supervised fine-tuning and DR staging on APTOS. Metrics evaluated include Accuracy (Acc), AUC ROC, qKappa, F1, and rDR Accuracy (rDR Acc). The top panel shows the results obtained when fine-tuning on the full APTOS dataset. In contrast, the bottom panel presents the average of 5 runs from the few-shot study on APTOS with 16 sample images per class and includes error bars representing SDs. Evaluation metrics are color-coded.

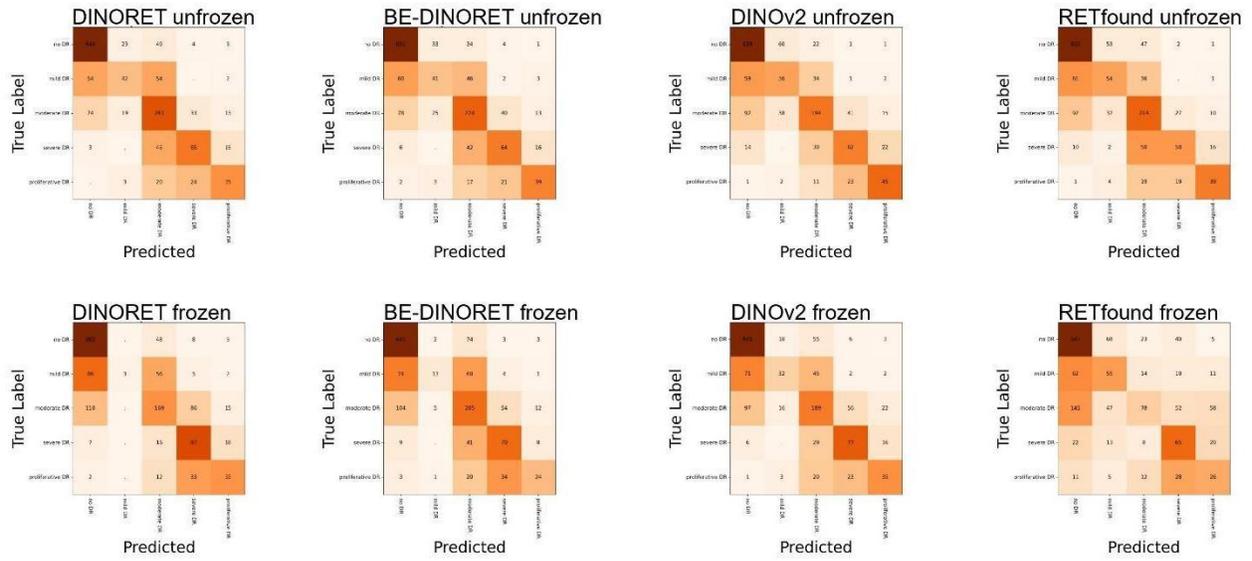

**Supplementary Figure 4: Confusion matrices for MSDFT experiments.** Models were trained, validated, and tested on a pooled dataset composed of all DR datasets. Each matrix's title specifies the model and backbone state. Ground-truth labels are shown on the vertical axis and predicted labels on the horizontal axis. Color intensity corresponds to the number of correct predictions, with darker colors indicating higher values and lighter colors indicating lower values.

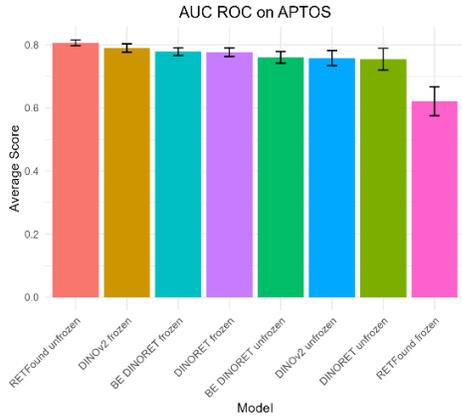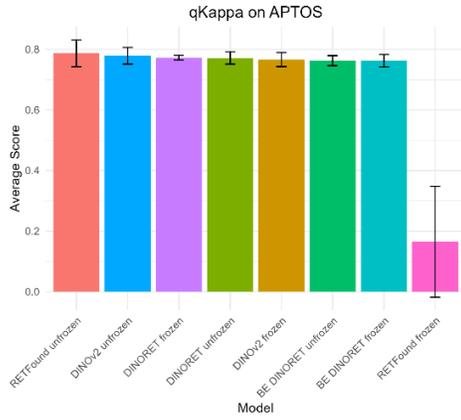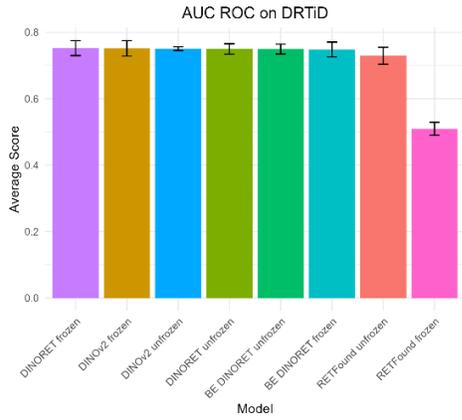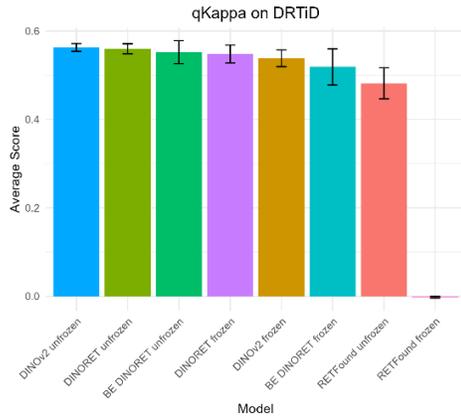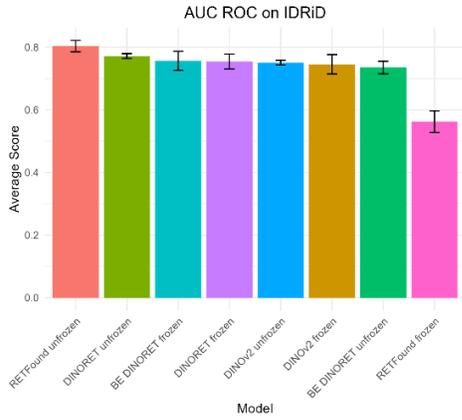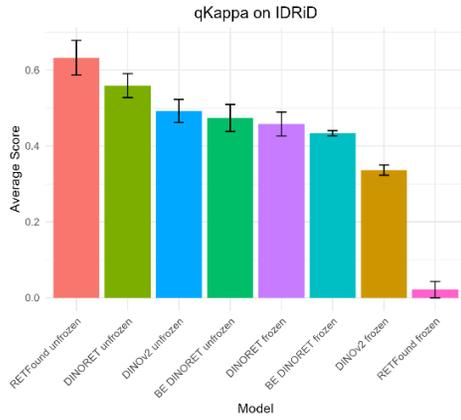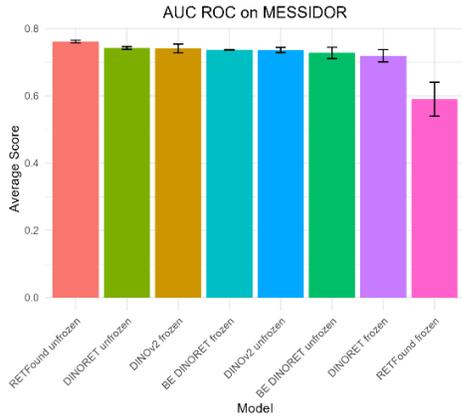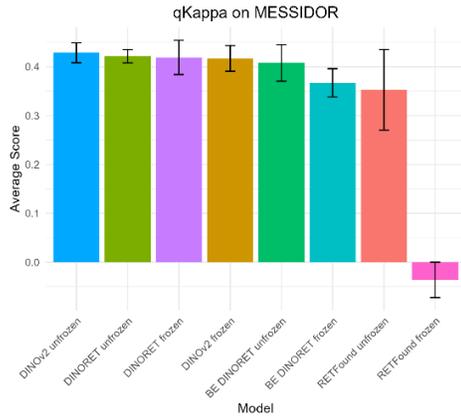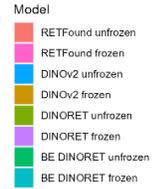

**Supplementary Figure 5: Average scores on a dataset for cross-evaluation experiments.** This figure presents the average AUC ROC and qKappa scores for all eight models tested on the four DR datasets during cross-evaluation experiments. Each bar represents the mean score of a model on a single test dataset, averaged across all three cross-evaluation experiments for a model. In each cross-evaluation experiment, models were trained on one dataset and tested on a separate dataset, resulting in three distinct test results for each model on a dataset. These three scores were averaged to obtain the mean score for each model on each test dataset. Error bars indicate the SEM. The x-axis lists the models, each depicted by a distinct color, and the y-axis displays the average scores a model achieves. Plot titles specify the metric (AUC ROC or qKappa) and the test dataset. Models are arranged in descending order of their average scores from left to right within each plot, with distinct colors to differentiate the models.

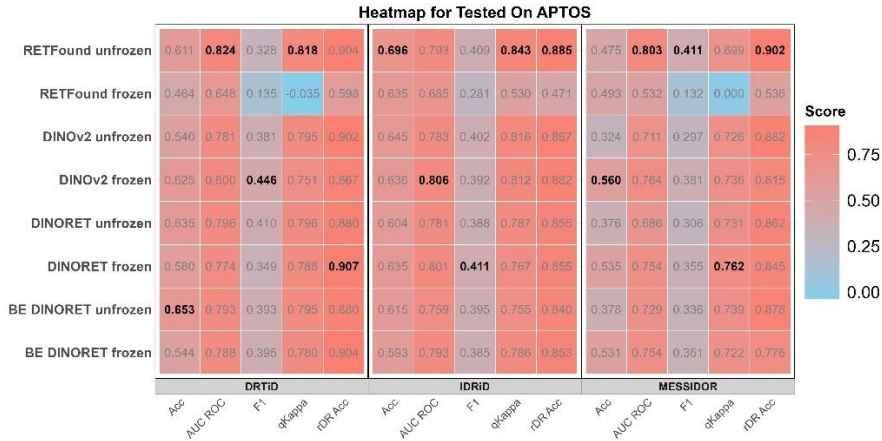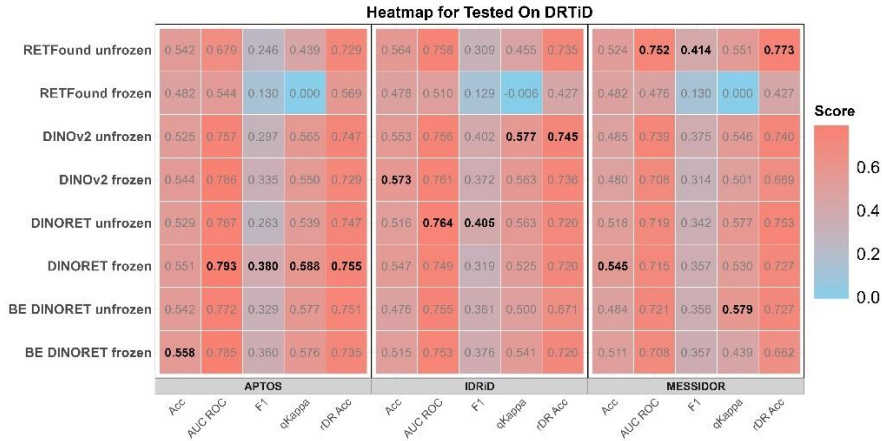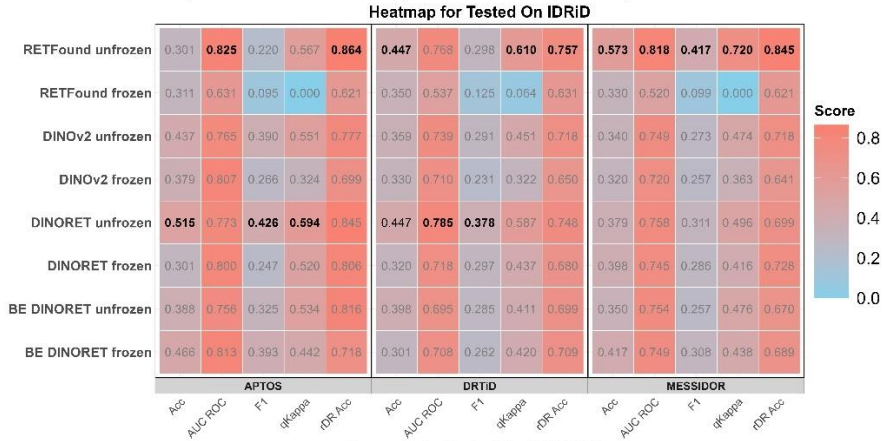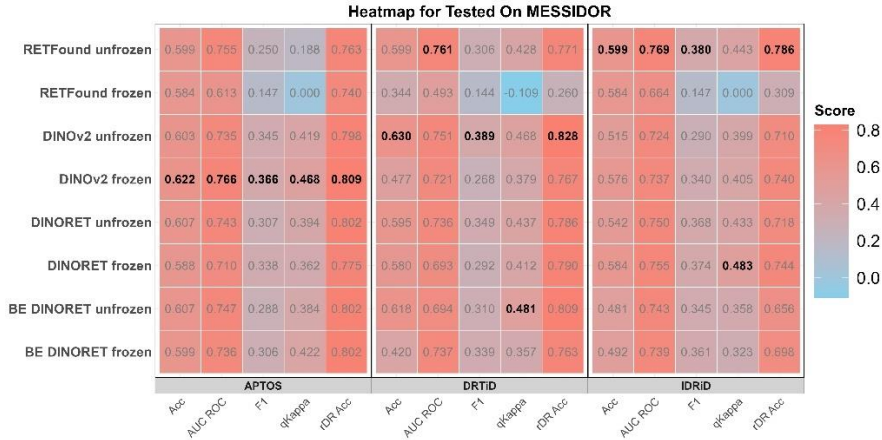

**Supplementary Figure 6: Cross-evaluation experiments and external validation.** This figure presents heatmaps for all cross-evaluation experiments performed in this study, separated by the four distinct test datasets. Each subplot title indicates the test dataset, with the datasets used for training separated along the x-axis and annotated in the gray bars. The models are listed on the y-axis, and the evaluation metrics are on the x-axis. The color gradients represent the score, ranging from light blue (lower scores) to red (higher scores), with the best model per metric (column) being indicated in bold.

# AUC ROC

## PAPILA
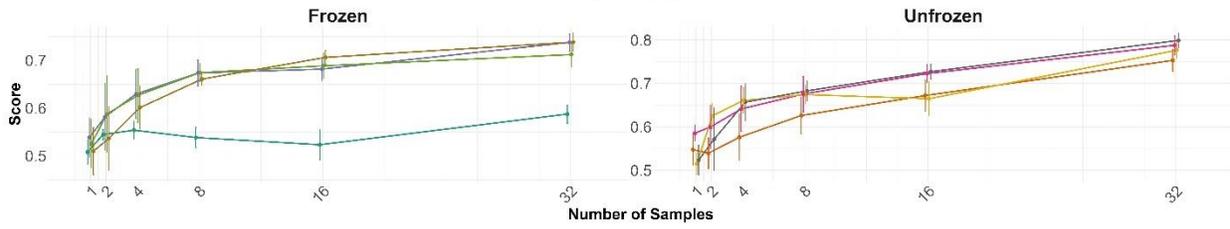

## DRTiD
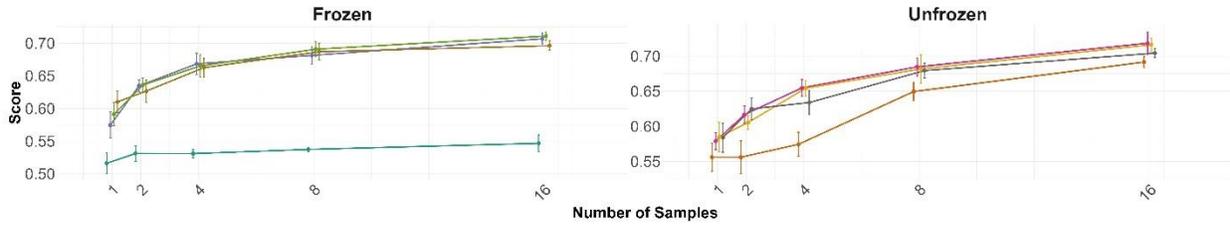

## MESSIDOR
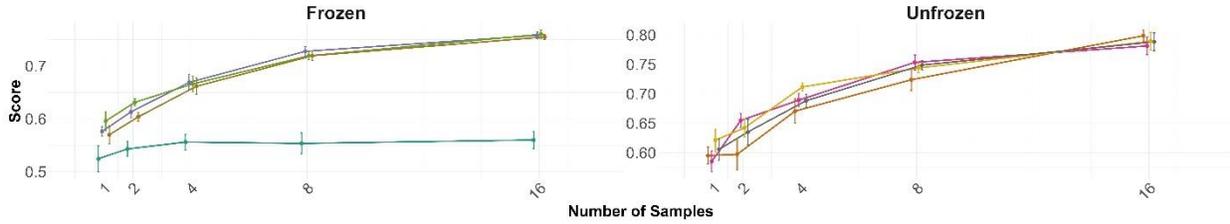

## APTOS
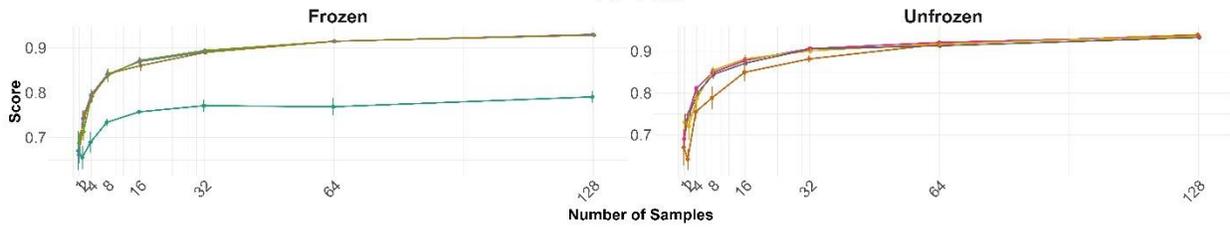

## IDRiD
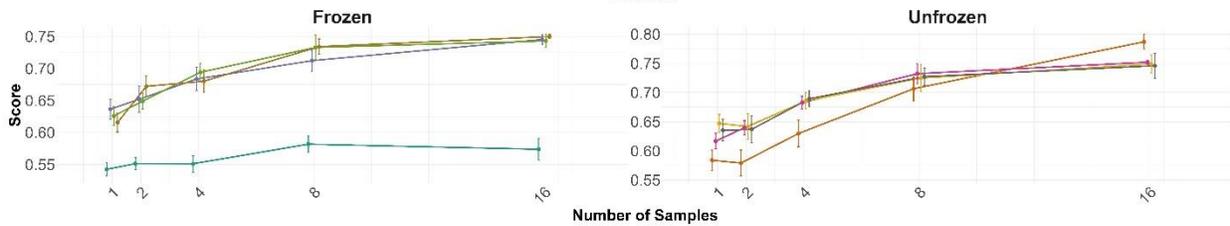

# qKappa

## PAPILA
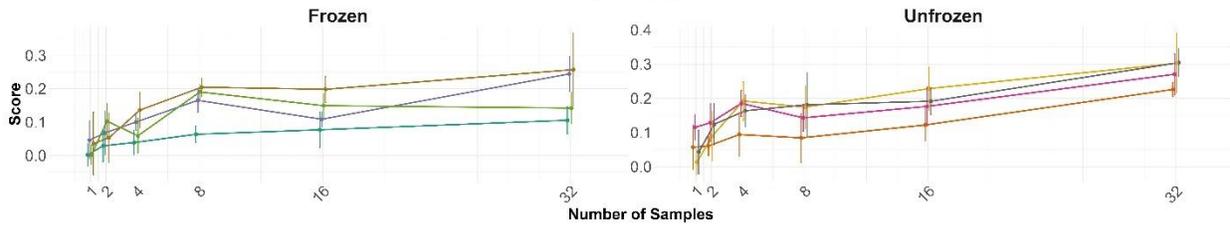

## DRTiD
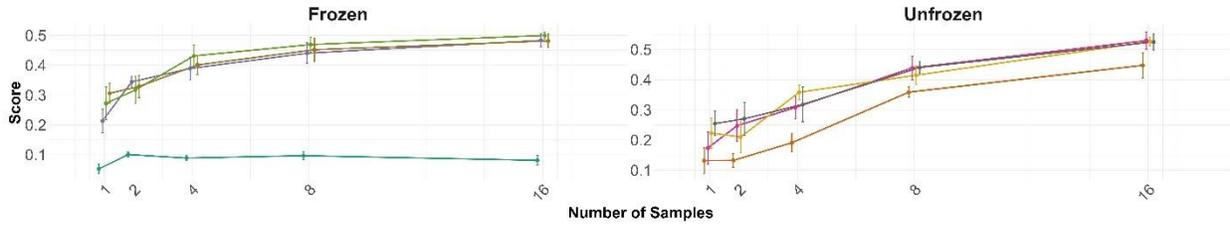

## MESSIDOR
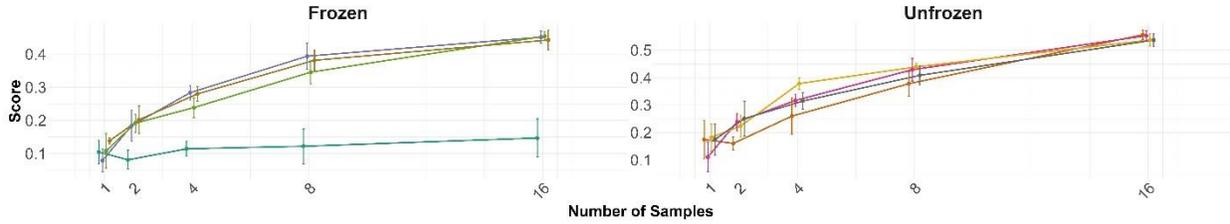

## APTOS
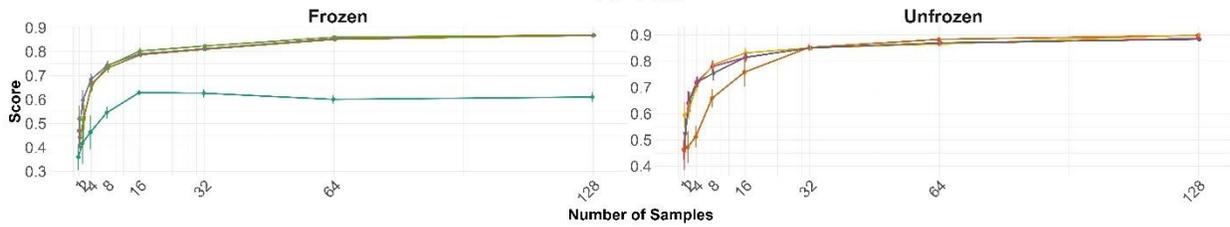

## IDRiD
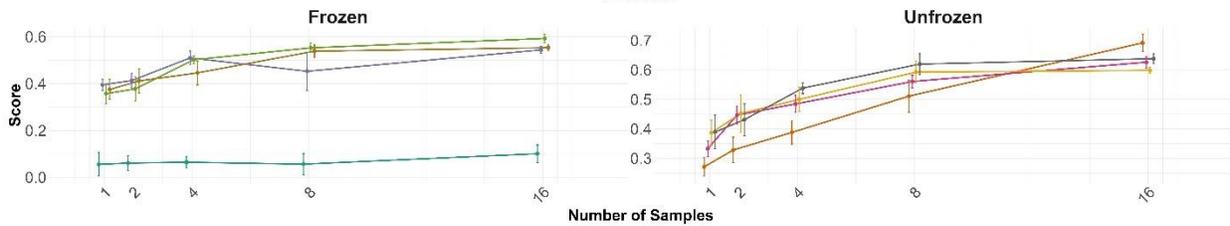

# Accuracy

## PAPILA
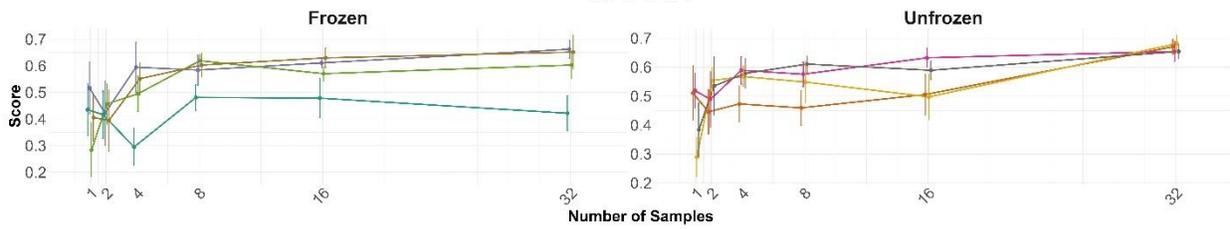

## DRTiD
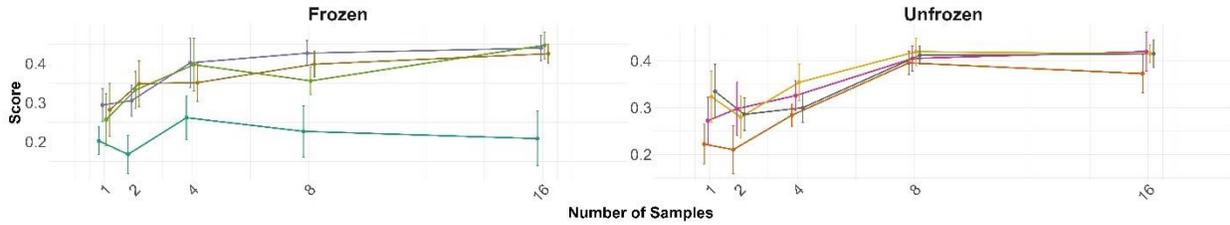

## MESSIDOR
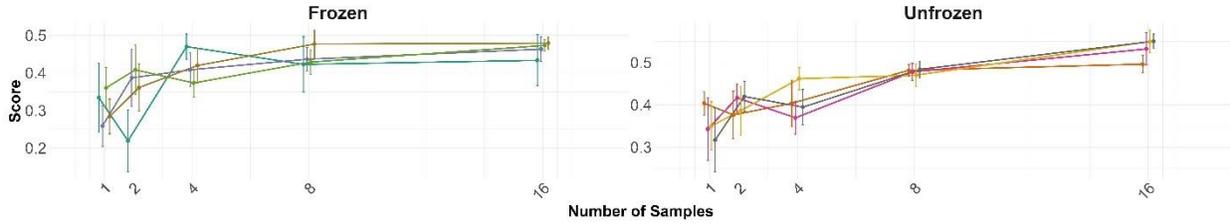

## APTOS
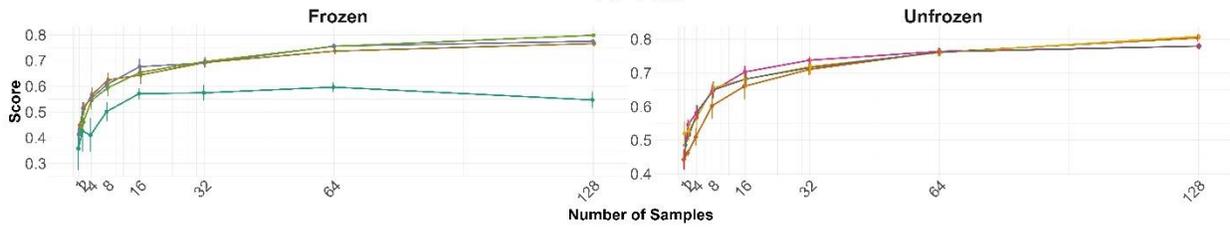

## IDRiD
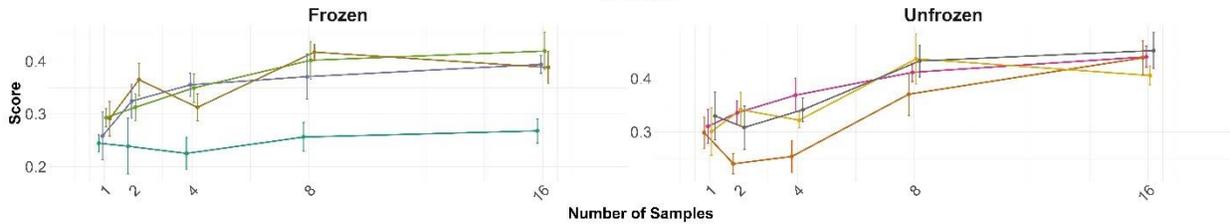

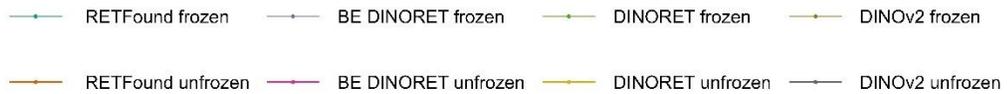

# F1
## PAPILA
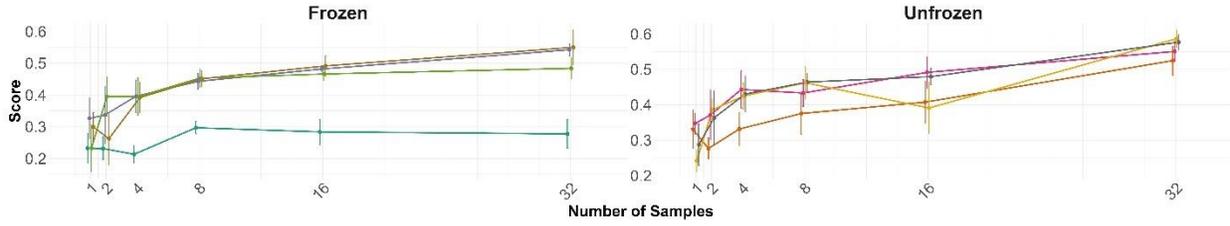

## DRTiD
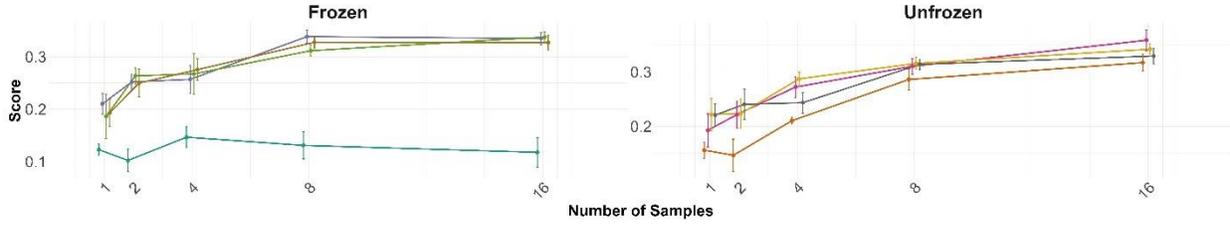

## MESSIDOR
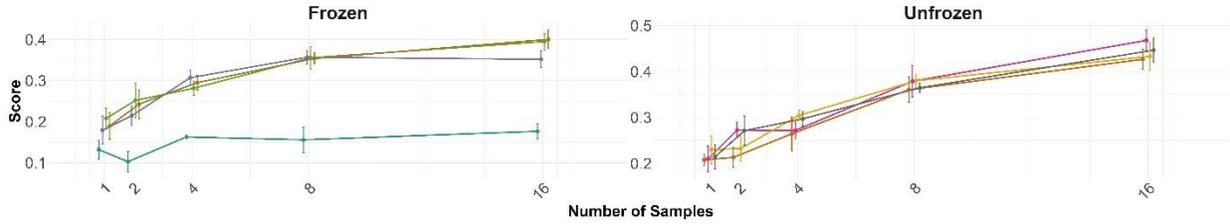

## APTOS
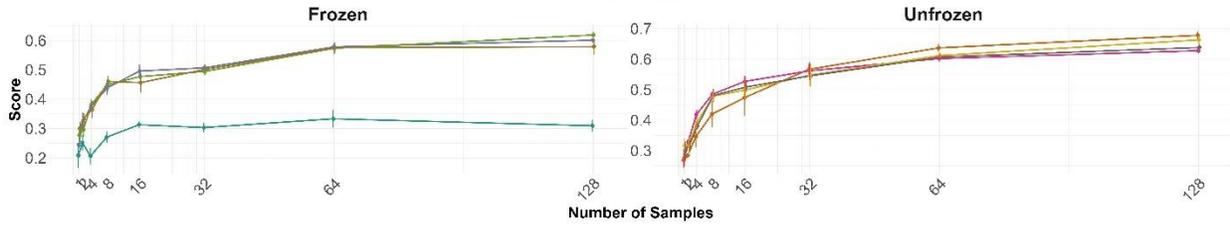

## IDRiD
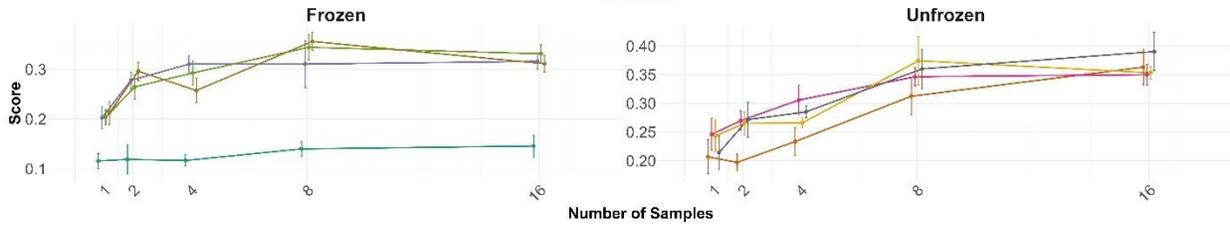

# rDR Accuracy
## PAPILA

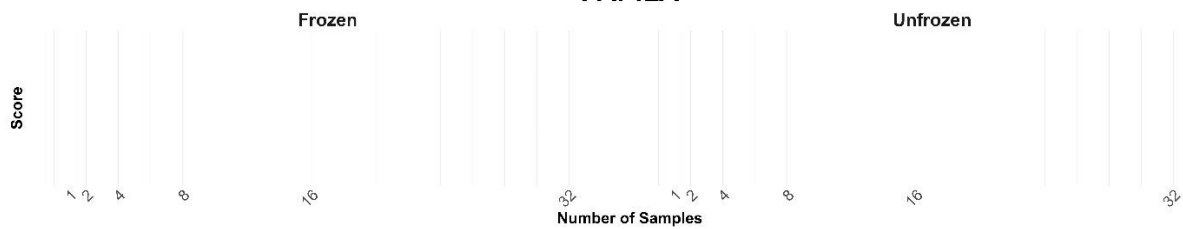

## DRTiD

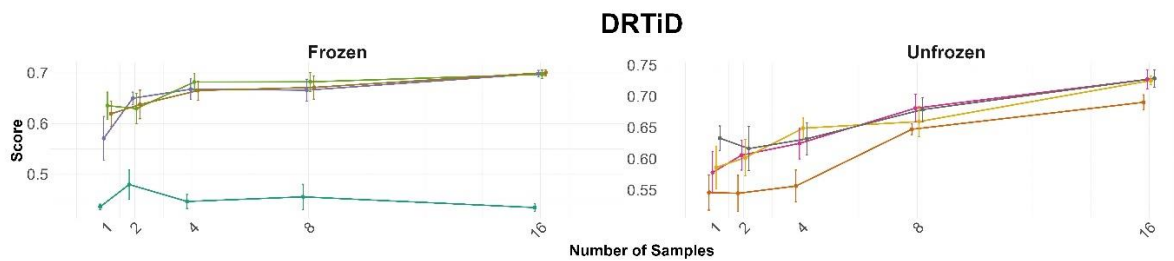

## MESSIDOR

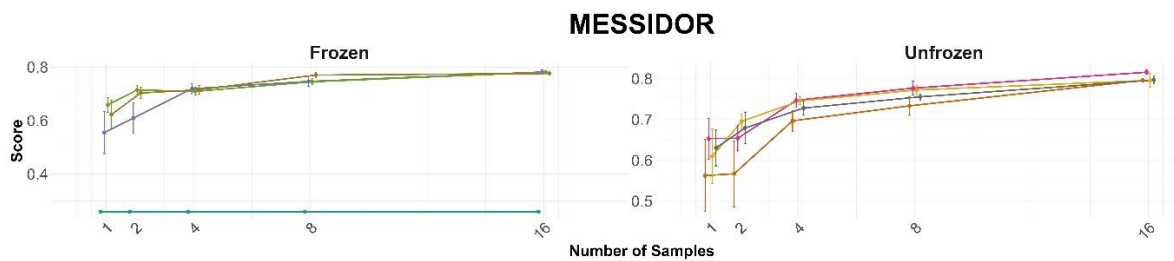

## APTOS

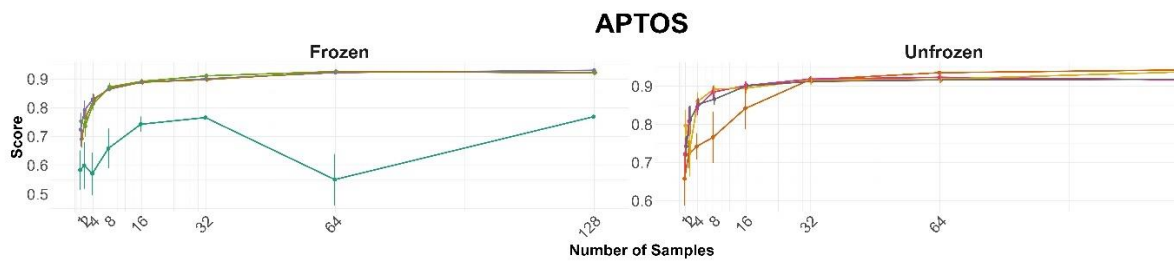

## IDRiD

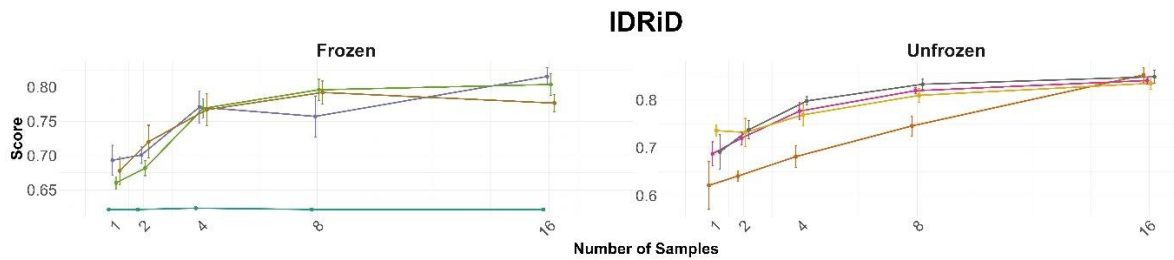

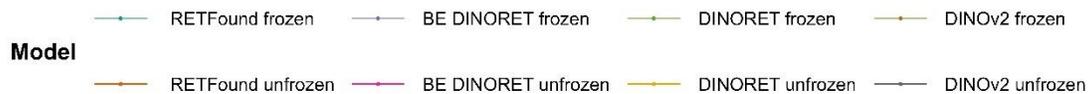

**Supplementary Figures 7-11: Data efficiency by evaluation metrics for few shot experiments.** The figures illustrate the results obtained from our few-shot experiments on all 5 datasets, for frozen and unfrozen BBs during supervised training. Each plot depicts the results for a specific evaluation metric, with each subplot representing a distinct dataset. The x-axis represents the number of sample images per class (few-shot), while the y-axis indicates the performance score for the respective metric. Models are color-coded, and error bars denote the SEM across 5 runs. The plot titles correspond to the evaluation metric, and subtitles refer to the datasets.

# 8. Appendix

## 8.1 Supplementary Material 1

### Datasets for Self-Supervised Post-Pretraining

**Eye Picture Archive Communication System (EYEPACS)**
The Kaggle-EYEPACS dataset is a large, open-access collection of CFPs for diabetic retinopathy (DR) research (1–3). This dataset, initially provided for a Kaggle competition, encompasses 88,702 macula-centered CFPs of diverse resolutions, obtained using various cameras across multiple locations (primarily in the USA). For our study, 88,699 images of the original dataset were used (three images appeared damaged), with 17,965 images being classified as ungradable by AutoMorph (4).

**Artificial Intelligence for RObust Glaucoma Screening (AIROGS)**
The Artificial Intelligence for RObust Glaucoma Screening (AIROGS) dataset was released as part of the International Symposium on Biomedical Imaging (ISBI) 2022 challenge program (5). 101,442 CFPs from 54,274 subjects (test subset) are publicly available and utilized in this study (except 175 damaged images). The dataset contains images classified as referable glaucoma (3.2%) and non-referable glaucoma (96.8%) and images were obtained using several devices (5). A total of 25,266 images were classified as ungradable by AutoMorph (4).

**DDR**
The DDR dataset comprises 13,673 CFPs sourced from 147 hospitals across 23 provinces in China. The images are derived from 9,598 patients, with a nearly even gender distribution of 48.23% male and 51.77% female and a mean age of 54 years (6). The dataset utilizes 42 types of fundus cameras, ensuring diverse imaging conditions. For our experiments, all images (except 69 damaged images) were utilized and filtered using AutoMorph (4), leading to the exclusion of 4,305 images deemed ungradable.

### Datasets for Supervised Fine-Tuning:

**Asia Pacific Tele-Ophthalmology Society (APTOS)**
The APTOS 2019 dataset, sourced from a Kaggle competition and provided by the Aravind Eye Hospital in India, features a comprehensive collection of 5,590 CFPs (7). These images, predominantly macula-centered, were captured by technicians in various rural areas of India, showcasing a diverse range of environmental and operational conditions. Images were graded by physicians according to the ICDR scale (8). For our research, we focused on the training subset of the APTOS dataset (APTOS public), which includes 3,662 CFPs, since ground-truth labels were not available for the test portion of the dataset (APTOS private). The training set from the APTOS competition was divided into distinct segments for training, testing, and validation, adhering to a ratio of 70:15:15 and maintaining an equal distribution of labels across all subsets. A total of 734 (19.77%) images were deemed ungradable by AutoMorph (4), but still included in this study.

**Methods to Evaluate Segmentation and Indexing Techniques in the Field of Retinal Ophthalmology (Messidor-2)**

The Messidor-2 dataset is an open-access dataset created to assist studies applying computer vision for diabetic retinopathy and kindly provided by the Messidor program partners (see https://www.adcis.net/en/third-party/messidor/) (9,10). The dataset contains a total of 1,748 macula centered CFPs from 874 examinations, across various resolutions. While the Messidor-2 dataset itself does not come with labels, annotations have been provided by third parties, including Google (11). For our study, we relied on the DR grades provided by Krause et.al., which were adjudicated by three retinal specialists (11). Additionally, information regarding the gradability of all images was also made available (four ungradable images), which we excluded for further experiments, leaving us with a final dataset of 1,744 images. A balanced, stratified split for train:validate:test of 70:15:15 was adopted and 52 images (2.97%) were deemed ungradable by AutoMorph (4).

**Indian Diabetic Retinopathy Image Dataset (IDRiD)**

The IDRiD dataset was originally made available as part of the "Diabetic Retinopathy: Segmentation and Grading Challenge," organized in conjunction with the IEEE International Symposium on Biomedical Imaging (ISBI-2018), Washington D.C. (12–14). The dataset comprises 516 macula-centered CFPs, all acquired using a Kowa VX-10 alpha camera at a resolution of 4,288 × 2,848 pixels. All photographs were collected at a single eye clinic in Nanded, Maharashtra, India (12). Ground truth labels for DR grades (ICDR scale) are provided together with the dataset and represent a consensus decision from 2 medical experts (8,12). All included images were deemed gradable and of good quality by the 2 medical experts. A defined test-set of 103 images for DR grading was proposed in the challenge and adopted for our study. The remaining 403 images were split into a test:validate set in a ratio of 82.5:17.5, maintaining equal class distribution. Overall this left us with a split of 66:14:20 for train:validate:test, with class imbalances between the train/validate and the test set. A total of 57 images (11%) were classified as ungradable by AutoMorph (4).

**The Diabetic Retinopathy Two-field image Dataset (DRTiD)**

DRTiD comprises 3,100 two-field (1 optic disc centered, 1 macula centered per eye) CFPs from 1,550 eyes, collected from the Shanghai Diabetic Eye Study between 2015 and 2017 (15). The images, captured with non-mydriatic retinal cameras with field of views ranging from 45° to 50°, feature resolutions between 1,444×1,444 and 3,058×3,058 pixels. Images were cleaned to meet high-quality standards and consistent fields of view. Ground truth labels were assigned by a panel of three experienced ophthalmologists using the ICDR scale, with intra-rater discrepancies resolved by an additional senior ophthalmologist (8,15). For our analysis, only the macula-centered images were utilized, resulting in a single image per eye. The defined test-set proposed by the authors (550 images) was adopted and the remaining 1,000 images were split into a class-balanced training and validation set adhering to a ratio of 82.5:17.5. Overall, we used a train:validate:test split of 54:11:35. According to AutoMorph, 508 images (33.66%) were of ungradable image quality (4). The dataset is accessible upon contacting the first author.

**PAPILA**

The PAPILA dataset includes bilateral CFPs from 244 patients, collected at the Hospital General Universitario Reina Sofía, Murcia, Spain (16). Ground truth labels were independently assigned to each eye by 2 physicians and based on a comprehensive clinical evaluation, with further verification through retrospective analysis of medical records. Thus, labels were not only assigned based on the presence of defining features in the CFPs but also included three classes: non-glaucomatous, suspected-glaucomatous, and glaucomatous (16). Overall, 488 CFPs were available and used in this study. A balanced, stratified split of 70:15:15 was used to obtain train:validate:test sets.

## 8.2 Supplementary Material 2

### Self-Supervised Post-pretraining Method

In our study, post-pretraining refers to updating the weights in the BB of the natural-domain pre-trained ViT DINOv2, whose weights and biases are initialized (17). During post-pretraining, weights across layers are updated using the DINOv2 SSL pipeline on CFPs (17). For DINORET, all blocks were unfrozen, updating weights across all blocks (Figure 1), while for BE-DINORET, the 12 original ViT blocks were frozen and only the parameters within duplicated blocks were changed during SSL on CFPs (Figure 1). The data used for post-pretraining consisted of the CFP-Large dataset (a combination of Kaggle-EYEPACS, AIROGS and DDR), collectively amounting to 203,570 images. These images are all used without ground-truth labels for SSL. To increase the number of images, all CFPs are split into patches (the amount depending on the model configuration, as depicted in Supplementary Figure 2, and as applied in similar research (18). Each patch (1-25 per CFP) is resized to 224 x 224 pixels. Finally the models are fine-tuned using a modified DINOv2 pipeline, which is a form of contrastive SSL (19), with an adapted code from the original DINOv2 repository (17,20). These adaptations include BE for BE models only (21) and a custom dataset (CFP-Large). Furthermore, the job submission methods are modified to accommodate our Simple Linux Utility for Research Management (SLURM) environment and aggregation steps are included in the training loop, allowing for a larger batch size and consideration of the available Video Random Access Memory (VRAM). Otherwise, the code is identical to the one in the public DINOv2 repository. For evaluation of post-pretraining approaches, each model is fine-tuned for several downstream tasks using supervised learning (SL), including DR staging on the full APTOS dataset and a few-shot-study on this dataset with 16 training images per class. Based on the performance on these downstream tasks, model evaluation, selection and hyperparameter tuning is performed. As suggested in previous studies, we disable the Koleo regularizer to accelerate and stabilize the training process at scale (18).

## 8.3 Supplementary Material 3

### Formulas for Evaluation Metrics

**F1 Score:**

Precision: Precision = TP / (TP + FP)

Recall: Recall = TP / (TP + FN)

F1 Score = 2 * (Precision * Recall) / (Precision + Recall)

Where:

TP = True Positives

FP = False Positives

FN = False Negatives

**Overall Accuracy (Acc):**

Accuracy: Accuracy = (TP + TN) / (TP + TN + FP + FN)

Where:

TP = True Positives

TN = True Negatives

FP = False Positives

FN = False Negatives

**Area Under the Receiver Operating Characteristic Curve (AUC ROC)**

AUC ROC=$\int_0^1 TPR(t)d(FPR(t))$

Where:

TPR($t$) = TPR at threshold $t$.

FPR($t$) = FPR at threshold $t$.

True Positive Rate (TPR): TPR (Sensitivity) = TP / (TP + FN)

False Positive Rate (FPR): FPR (1 - Specificity) = FP / (FP + TN)

TP = True Positives

FN = False Negatives

FP = False Positives

TN = True Negatives

**Quadratic Weighted Kappa (qKappa)**

Weight matrix element: $w_{ij} = (i - j)^2 / (N - 1)^2$

Expected frequency of ratings: $E_{ij} = A_i * B_j / T$

Observed weighted sum: $O_w = \Sigma \Sigma (w_{ij} * O_{ij})$

Expected weighted sum: $E_w = \Sigma \Sigma (w_{ij} * E_{ij})$

qKappa: $qKappa = 1 - (O_w / E_w)$

Where:

$w_{ij}$ = Weight matrix element for ratings i and j

$O_{ij}$ = Observed frequency of ratings i and j

$E_{ij}$ = Expected frequency of ratings i and j

$A_i$ = Number of ratings i by Rater A (human, ground-truth)

$B_j$ = Number of ratings j by Rater B (ML model, interfered label)

T = Total number of ratings

N = Number of possible ratings

**k-Nearest Neighbors (kNN) Score:**

1. Extract embeddings generated by the ViT for each image.

2. For each image, find the k nearest neighbors in the feature space.

3. Assign the label that is most common among the k neighbors to the image.

4. Calculate the accuracy as the proportion of correctly classified validation images.

**Binary Classification Score for DR:**

Non-referable DR: Stage 0 and Stage 1

Referable DR: Stage 2, Stage 3, and Stage 4

**Scaled Loss:**

The scaling factor f_{Class i} for the loss contribution of each class i is defined as:

f_{Class i} = N_{Train, Total} / (N_{Train, Class i} * n_{Classes})

Where:

N_{Train, Total}: Total number of training samples.

N_{Train, Class i}: Number of training samples within the respective Class i.

## Training Hyperparameters:

During training of the DINOv2 based models, we use the AdamW optimizer with weight decay and a learning rate schedule with linear warmup and cosine annealing (22). Data augmentation methods for the DINOv2-based models include random resizing and cropping, colorjitter, dropout layers, dropout paths, random horizontal and vertical flipping of images, random rotation of images, and random sharpness adjustments (23). As some datasets are imbalanced, we apply a scaling to the loss that is dependent on the number of sample images for a specific class in the dataset, as shown below. For RETFound, augmentations are performed as specified in their source code (24). The final model is chosen based on the best qKappa checkpoint on the respective validation set.

21. Wu C, Gan Y, Ge Y, Lu Z, Wang J, Feng Y, u. a. LLaMA Pro: Progressive LLaMA with Block Expansion [Internet]. arXiv; 2024 [zitiert 26. Januar 2024]. Verfügbar unter: http://arxiv.org/abs/2401.02415

22. Loshchilov I, Hutter F. Decoupled Weight Decay Regularization. In: International Conference on Learning Representations [Internet]. 2017. Verfügbar unter: https://api.semanticscholar.org/CorpusID:53592270

23. Ayzenberg L, Giryes R, Greenspan H. DINOv2 based Self Supervised Learning For Few Shot Medical Image Segmentation [Internet]. arXiv; 2024 [zitiert 23. Mai 2024]. Verfügbar unter: http://arxiv.org/abs/2403.03273

24. Zhou Y, Chia MA, Wagner SK, Ayhan MS, Williamson DJ, Struyven RR, u. a. A foundation model for generalizable disease detection from retinal images. Nature. 5. Oktober 2023;622(7981):156–63.

## 8.4 Supplementary Results 1

### Supervised Fine-tuning Parameter Studies:

In these studies, we evaluated if using the average patch embeddings instead of the [CLS] token or both concatenated for image classification improves DR staging with DINORET, BE DINORET, or DINOv2. We performed experiments similar to the cross-evaluation experiments. In short, models were trained on all four DR datasets individually, with both frozen and unfrozen BBs, and evaluated on the test sets of each DR dataset, resulting in 16 total experiments per model. Subsequently, we compared DR staging performance when using the [CLS] token, the average patch embeddings, or both concatenated. As illustrated in Supplementary Table ST1.1, using the [CLS] token for DR staging resulted in a better qKappa score than the other embeddings in 6/16 tasks for unfrozen DINOv2, 3/16 for unfrozen DINORET, and 8/16 for unfrozen BE DINORET. Supplementary Figure SR1.1 depicts all results obtained from these experiments. Additionally, we performed a few-shot study on all DR datasets for the DINOv2 model, where runs were performed in quintuplicate and until the sample count on the training dataset could not be increased further, as illustrated in Supplementary Figure SR1.2. For unfrozen DINOv2, using the patch embeddings resulted in a significantly higher qKappa score on APTOS with 32 training images per class, compared to the [CLS] token ($p = 0.049$) and the concatenated embeddings ($p = 0.028$, ANOVA with Tukey HSD), and a significantly higher qKappa score on IDRiD compared to the [CLS] token with 16 training images per class ($p = 0.034$, ANOVA with Tukey HSD). All other differences lacked significance, as shown in Supplementary Data 2. Finally, we investigated whether the role of using a distance weighted, scaled cross entropy loss (CEL) function would improve DR staging with fine-tuned models. All 16 cross-evaluation comparisons were repeated with DINOv2, using both the vanilla CEL and the scaled CEL. The median score was higher for the scaled CEL for all evaluation metrics, except for qKappa, as depicted in Supplementary Figure SR1.3.

| Model | BB | Metric | [CLS] | Concatenated | Patch | Total Experiments |
|---|---|---|---|---|---|---|

| Model | BB state | Metric | CLS | Patch | Both | Total |
|---|---|---|---|---|---|---|
| BE DINORET frozen | frozen | AUC ROC | 5 | 8 | 3 | 16 |
| BE DINORET unfrozen | unfrozen | AUC ROC | 3 | 8 | 5 | 16 |
| DINORET frozen | frozen | AUC ROC | 5 | 1 | 10 | 16 |
| DINORET unfrozen | unfrozen | AUC ROC | 5 | 3 | 8 | 16 |
| DINOv2 frozen | frozen | AUC ROC | 7 | 7 | 2 | 16 |
| DINOv2 unfrozen | unfrozen | AUC ROC | 7 | 4 | 5 | 16 |
| BE DINORET frozen | frozen | qKappa | 6 | 4 | 6 | 16 |
| BE DINORET unfrozen | unfrozen | qKappa | 8 | 6 | 2 | 16 |
| DINORET frozen | frozen | qKappa | 7 | 1 | 8 | 16 |
| DINORET unfrozen | unfrozen | qKappa | 3 | 2 | 11 | 16 |
| DINOv2 frozen | frozen | qKappa | 7 | 3 | 6 | 16 |
| DINOv2 unfrozen | unfrozen | qKappa | 6 | 7 | 3 | 16 |

**Supplementary Table ST1.1: Embedding strategies used for classification influence performance on downstream tasks.** For each model, 16 experiments were performed, by individually training and testing on all 4 DR datasets, resulting in 16 total evaluations across which embedding strategies can be compared. Subsequently, we compared the performance of DR staging, when using the [CLS] token, the average patch embeddings, or both concatenated for classification. The table indicates models, their respective BB states and the number of experiments in which the [CLS] token, the batch embeddings or both concatenated achieved the best score for a given metric.

Heatmap for Tested On APTOS

Heatmap for Tested On DRTiD

Heatmap for Tested On IDRiD

Heatmap for Tested On MESSIDOR

**Supplementary Figure SR1.1: Embedding strategies and DR staging performance.** This figure presents heatmaps for all experiments performed when comparing embeddings for DR classification in this study and separated for four distinct test datasets. Each subplot title indicates the test dataset, with the datasets used for training separated along the x-axis and annotated in the gray bars. The models with their BB state and the embedding used for classification are listed on the y-axis, and the evaluation metrics are on the x-axis. The color gradients represent the score, ranging from light blue (lower scores) to red (higher scores), with the best model per metric (column) being indicated in bold.

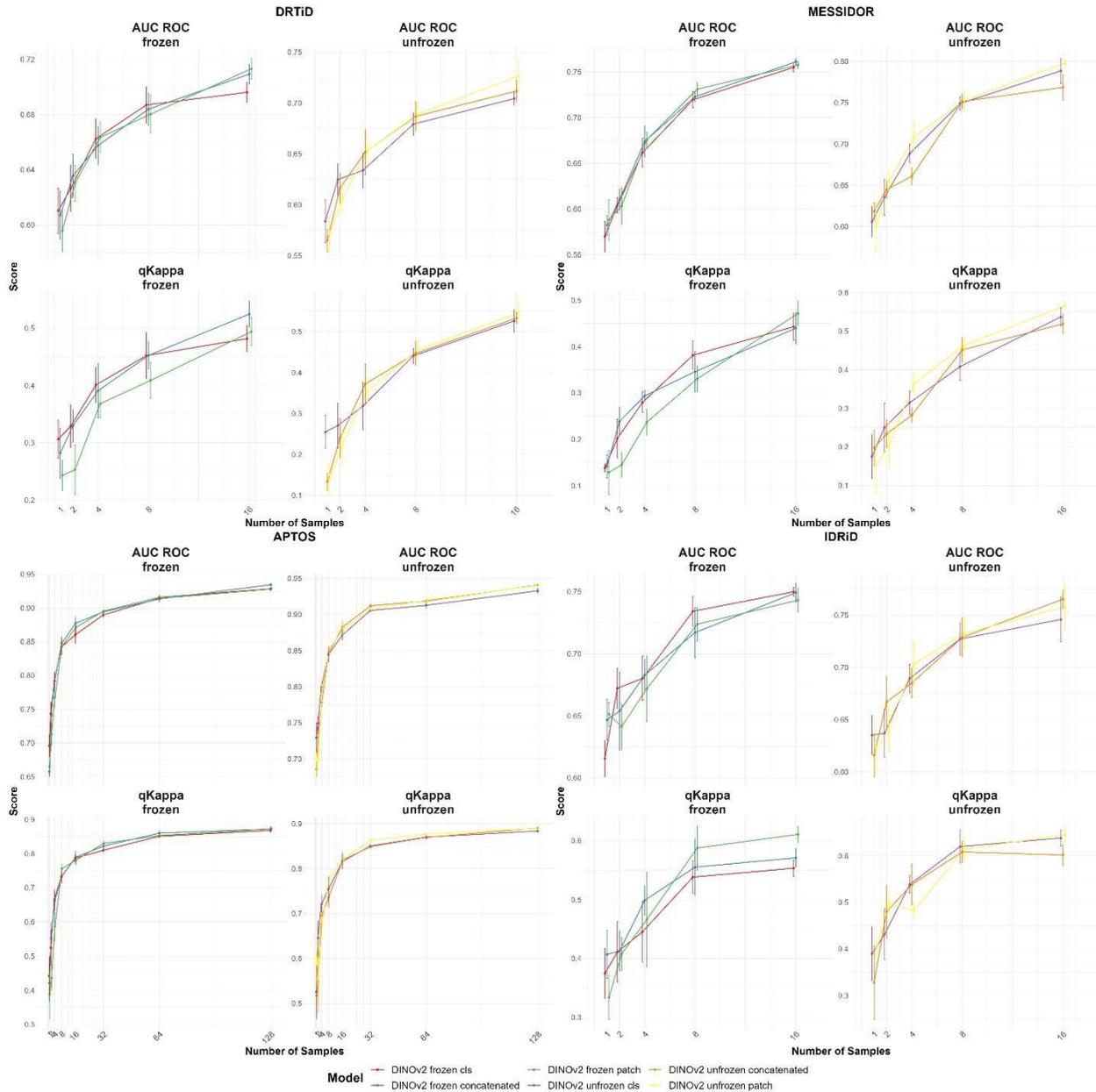

**Supplementary Figure SR1.2: Embedding strategies and data efficiency for DR staging.** This figure illustrates the results of a few-shot learning study performed on the four DR datasets. The subplots depict the performance of DINOv2l trained with a varying number of sample images per class, with separate evaluations for using the [CLS] token, the average patch embeddings, and both concatenated. The x-axis represents the number of images per class

used for training, while the y-axis depicts the qKappa and AUC ROC scores. The data is further divided into facets based on the evaluation metric (AUC ROC or qKappa) and the BB state.

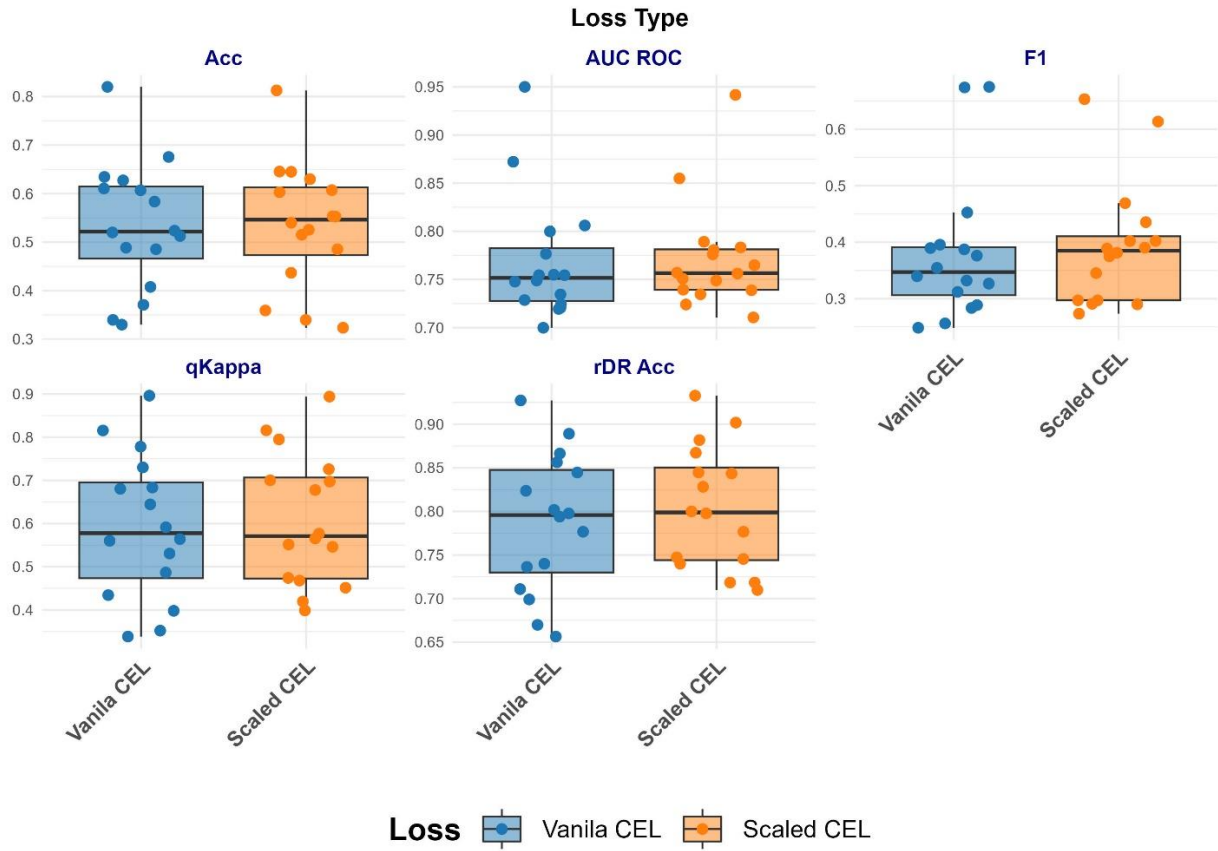

**Supplementary Figure SR1.3: A distance weighted, scaled CEL might improve DR staging.** Comparison of Vanilla CEL and Scaled CEL Losses for DR Staging. This figure shows the evaluation metrics (Accuracy, AUC ROC, qKappa, F1 Score, and rDR Accuracy) for DR staging, using the DINOv2 model and comparing Vanilla CEL and Scaled CEL losses during supervised fine-tuning. 16 tasks were evaluated, by training and testing DINOv2 on all possible combinations of the 4 DR datasets, once with a frozen and an unfrozen BB. The central line in each box indicates the median score, while the box itself delineates the interquartile range from the 25th to the 75th percentile. Whiskers extend from the boxes to 1.5 times the interquartile range, marking the range of typical data points, and individual scores are shown as dots, jittered to prevent overlap and enhance clarity.

# 8.5 Supplementary Results 2

## Cross-Evaluation:

As shown in Supplementary Table SR2.1, RETFound ranked amongst the best 2 models in 75% (out of 12 cross-evaluation experiments) for AUC ROC and 50% for qKappa (always unfrozen, never frozen). DINORET ranked amongst the best 2 models in 50% for AUC ROC (25% frozen, 25% unfrozen) and 75% for qKappa (25% frozen, 50% unfrozen). DINOv2 ranked amongst the best 2 models in 66.6% of cross-evaluations for AUC ROC (50% frozen, 16.6% unfrozen) and 33.3% for qKappa (25% unfrozen, 8.3 percent frozen). The BE DINORET model only ranked amongst the best two models in 8.3% for AUC ROC (always frozen) and 41.6% for qKappa (33.3% unfrozen, 8.3% frozen).

| Model | AUC ROC | | | qKappa | | |
| --- | --- | --- | --- | --- | --- | --- |
| | Frozen [%] | Unfrozen [%] | Total [%] | Frozen [%] | Unfrozen [%] | Total [%] |
| BE DINORET | 8.334 | 0 | 8.334 | 8.334 | 33.334 | 41.667 |
| DINORET | 25 | 25 | 50 | 25 | 50 | 75 |
| DINOv2 | 50 | 16.667 | 66.667 | 8.334 | 25 | 33.334 |
| RETFound | 0 | 75 | 75 | 0 | 50 | 50 |

**Supplementary Table ST2.1: Top 2 performance out of 12 cross-evaluation experiments.** This table displays the percentage cross-evaluations for which a model ranked amongst the best two models. Each experiment involved training the models on one DR dataset and testing them on a distinct dataset, resulting in 12 total cross-evaluations. For each model, the columns under AUC ROC and qKappa indicate the percentage of times a model ranked among the best 2 models when using frozen and unfrozen backbones, as well as the total percentage across both backbone states. Eight models were evaluated in total, obtained by fine-tuning RETFound, DINOv2, DINORET, and BE DINORET with frozen and unfrozen BBs.

# 8.6 Supplementary Results 3

## Unfrozen Fine-Tuning and Data Efficiency

For a pooled analysis, a hierarchical LMM was fit across all runs on all datasets, in the range approximately linearly increasing with the few-shot sample count (8-32). For both qKappa and AUC ROC, models were significant predictors of score outcomes (p = < 0.01, LRT). Post-hoc pairwise comparisons of models were conducted using estimated marginal means (EMMeans) with Kenward-Roger adjustment and compared via a Tukey HSD test. Unmodified DINOv2 significantly outperformed RETFound for AUC ROC and qKappa, DINORET and BE DINORET also outperformed RETFound, but differences were only significant for AUC ROC (BE DINORET) and qKappa (DINORET) individually, as shown in Supplementary Data 6. Differences between BE DINORET, DINORET, and unmodified DINOv2 were all insignificant. Supplementary Figures SR3.1 and SR3.2 present diagnostics from the model fits and Supplementary Figure SR3.3 displays the fitted average AUC ROC and qKappa scores for each model at a given sample count.

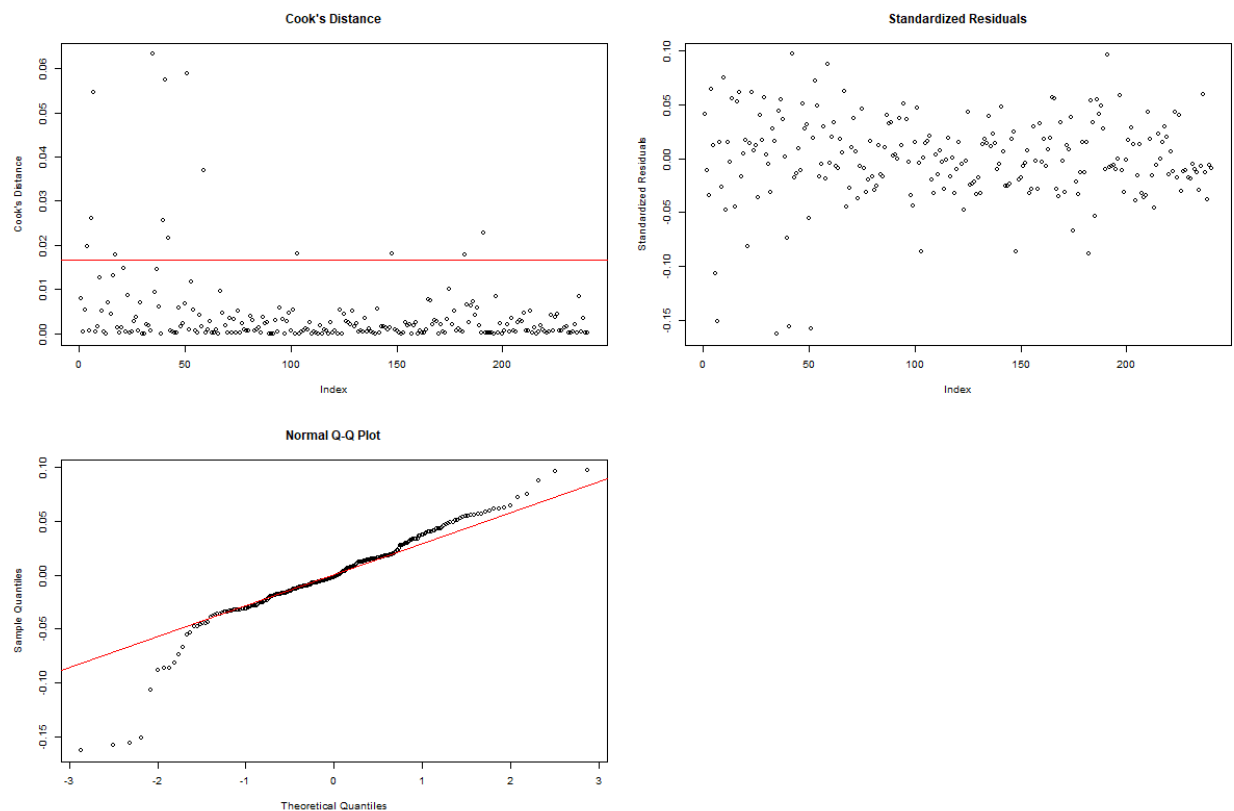

**Supplementary Figure SR3.1: Diagnostic plots from fitted linear mixed models with AUC ROC as the dependent variable.** The figure illustrates the diagnostic plots for the assessment of model fit, incorporating Cook's distance, standardized residuals, and a quantile-quantile (Q-Q) plot for the linear mixed-effects model examining the interaction effects of Few-Shot training counts and ViT model on AUC ROC. Cook's distance is plotted to identify

influential cases that could potentially distort the model estimations, with a reference line indicating a threshold of significant influence set at 4/n, where n is the number of observations. The standardized residuals plot assesses the homogeneity of variance and outliers, with horizontal lines at ±2 indicating typical bounds for standard deviations. The Q-Q plot assesses the normality of the residuals, providing a visual comparison between the observed residuals and those expected under a normal distribution, with deviations from the reference line suggesting departures from normality.

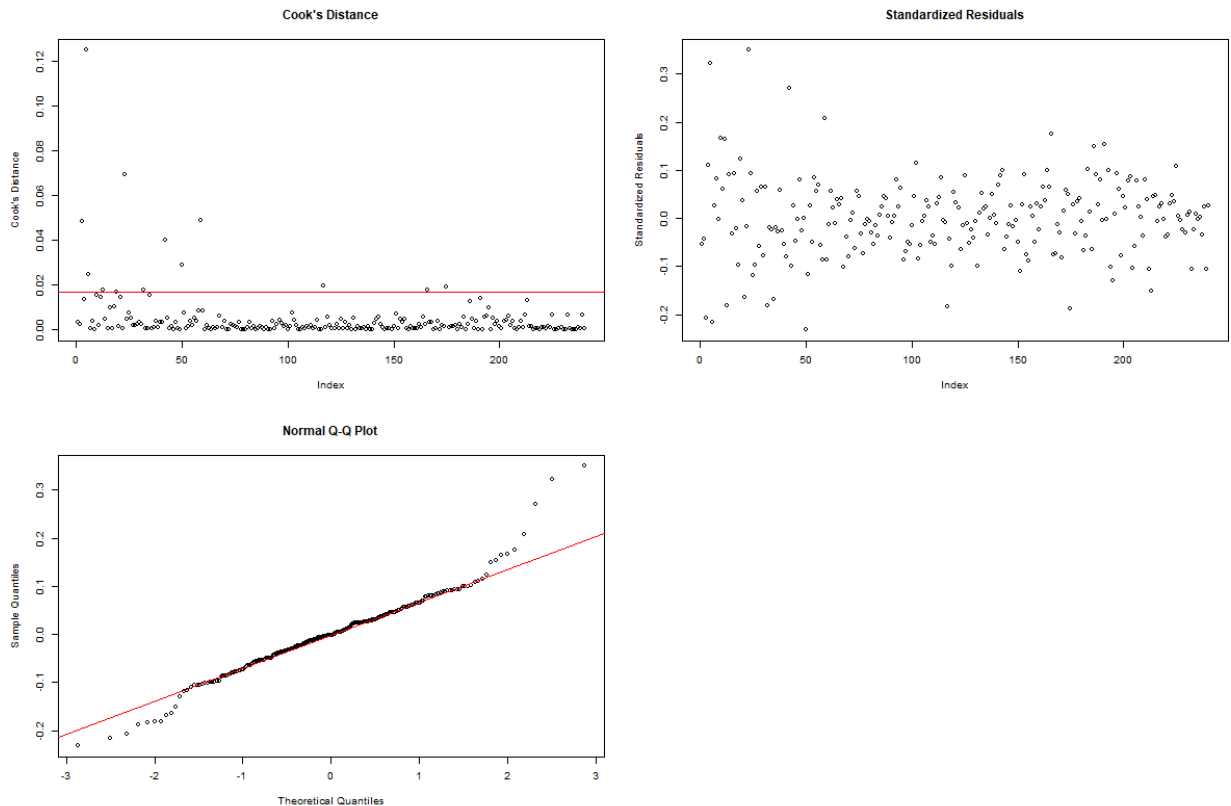

**Supplementary Figure SR3.2: Diagnostic plots from fitted linear mixed models with qKappa as the dependent variable.** The figure illustrates the diagnostic plots for the assessment of model fit, incorporating Cook's distance, standardized residuals, and a quantile-quantile (Q-Q) plot for the linear mixed-effects model examining the interaction effects of Few-Shot training counts and ViT model on qKappa.

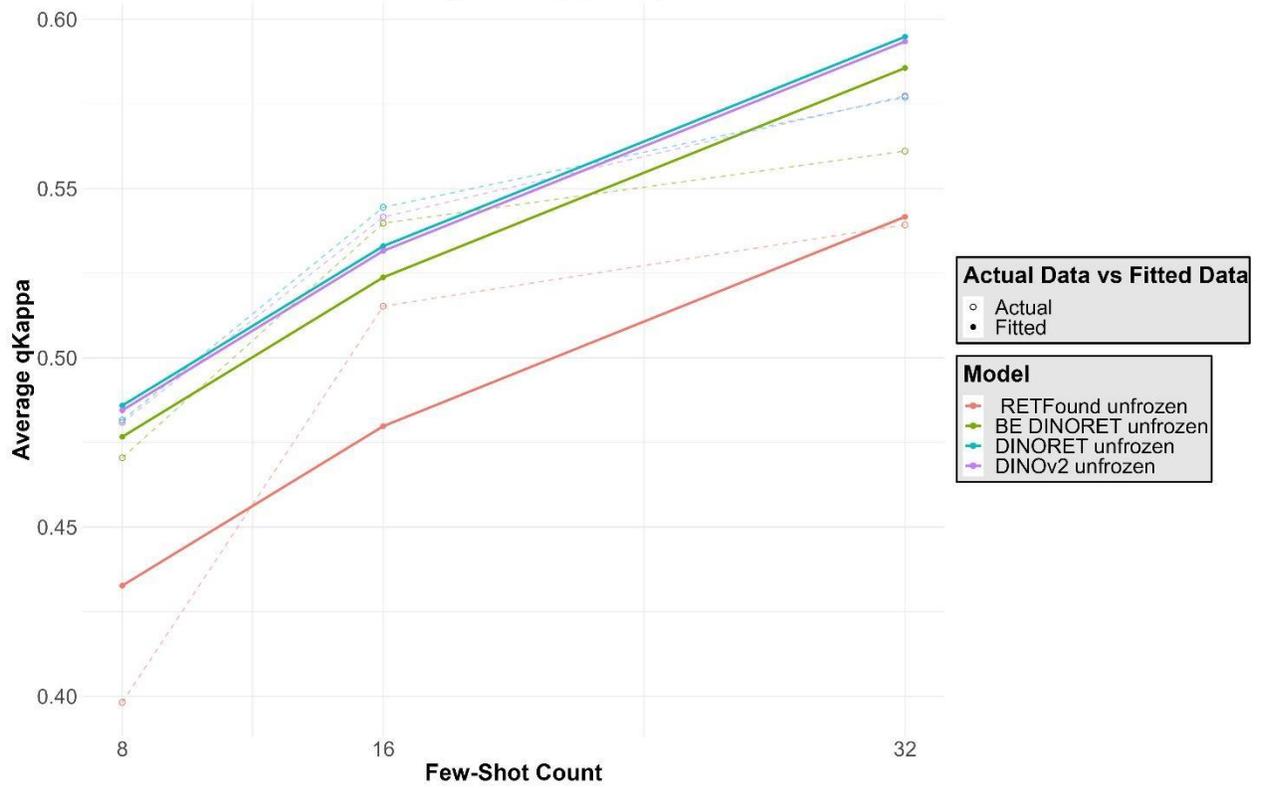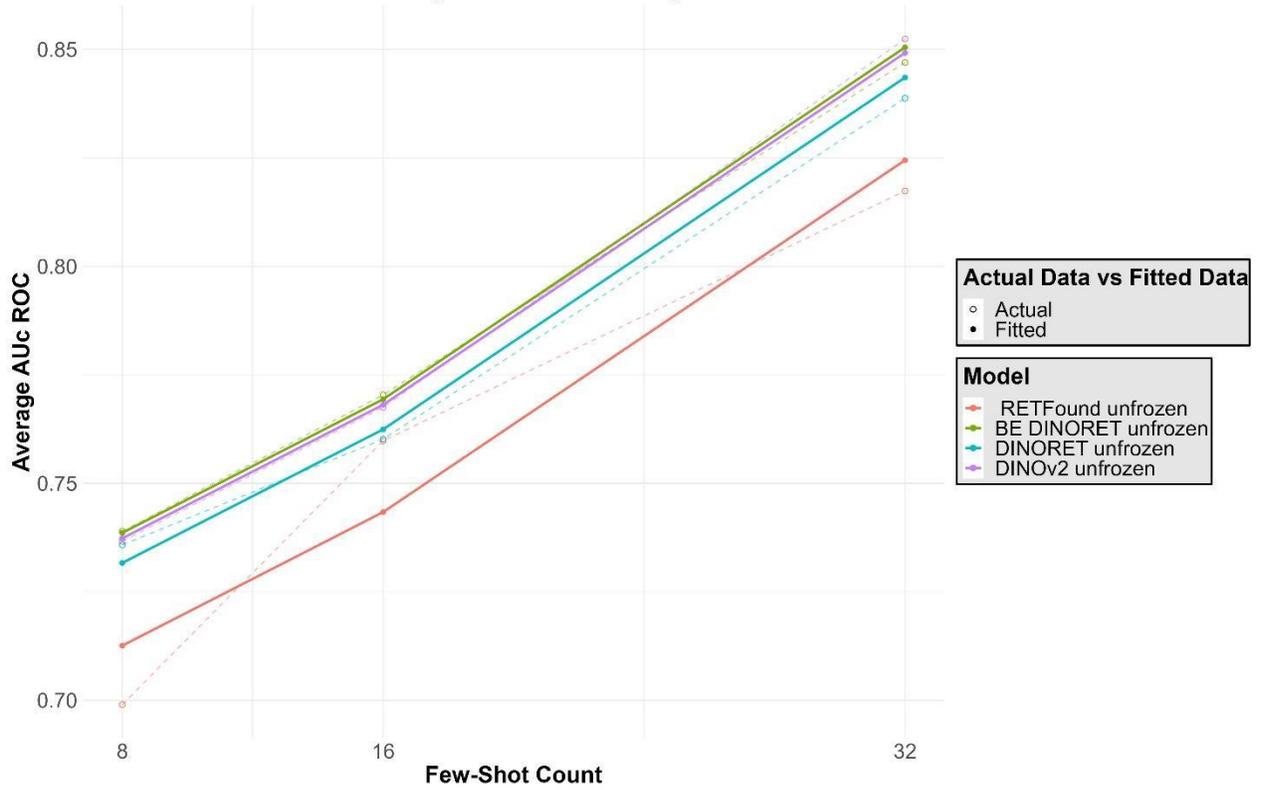

**Supplementary Figure SR3.3: Fitted averages for DR staging performance during few-shot learning.** This figure presents a comparative visualization between the average actual AUC ROC (top) and qKappa (bottom) values and those predicted by a linear mixed-effects model across Few-Shot training counts ranging from 8-32 per sample class. Each ViT model is differentiated by color. Solid lines with filled markers represent the predicted averages, highlighting the approximation by the mixed-effects model within the observed data range. In contrast, dashed lines with hollow markers delineate the actual observed averages. The plot distinctly allows for the assessment of the model's predictive performance, particularly focusing on the Few-Shot counts at 8, 16, and 32, thus enabling an evaluation of the model's utility in extrapolating training effectiveness.

# 9. Computational Resources

All experiments shown in this paper were run on a NVIDIA A10 cloud GPU with a 30 core CPU. The total training and evaluation time was approximately 1.5 weeks. This does not include the computation required for development.